\documentclass[11pt]{article}

\usepackage[preprint]{acl}

\usepackage{times}
\usepackage{latexsym}

\usepackage[T1]{fontenc}

\usepackage[utf8]{inputenc}

\usepackage{microtype}

\usepackage{inconsolata}

\usepackage{graphicx}

\usepackage{amsmath}

\newcommand{\medqa}{\textsc{MedQA}}
\newcommand{\imedqa}{i\textsc{MedQA}}
\newcommand{\craftmd}{\textsc{CraftMD}}
\newcommand{\icraftmd}{i\textsc{CraftMD}}
\newcommand{\mediq}{\textsc{MediQ}}

\usepackage{listings}
\usepackage{xcolor}
\usepackage{multirow}
\usepackage{makecell}

\definecolor{codebg}{RGB}{247,247,247}

\lstdefinestyle{appendixpython}{
  language=Python,
  basicstyle=\ttfamily\fontsize{6}{6.6}\selectfont,
  breaklines=true,
  breakatwhitespace=false,
  columns=flexible,
  keepspaces=true,
  showstringspaces=false,
  backgroundcolor=\color{codebg},
  frame=single,
  framerule=0pt,
  framesep=3pt,
  linewidth=\columnwidth,
  xleftmargin=0pt,
  xrightmargin=0pt,
  aboveskip=4pt,
  belowskip=4pt,
  resetmargins=true,
  tabsize=2,
  postbreak=\mbox{\textcolor{gray}{$\hookrightarrow$}\space}
}

\newcommand{\appparagraph}[1]{%
  \paragraph{#1}\mbox{}\par\nobreak\vspace{2pt}%
}

\title{LLM-Guided Evolution for Medical Decision Pipelines}

\author{
 \textbf{Ivan Sviridov\textsuperscript{1,*}},
 \textbf{Artem Oskin\textsuperscript{2}},
 \textbf{Ivan Panin\textsuperscript{2}},
 \\
 \textbf{Iaroslav Bespalov\textsuperscript{2}},
 \textbf{Dmitry Dylov\textsuperscript{2}},
 \textbf{Ivan Oseledets\textsuperscript{2}},
 \textbf{Aleksandr Nesterov\textsuperscript{2,*}}
\\
\\
 \textsuperscript{1}Sber AI Lab,
 \textsuperscript{2}AIRI
\\
 \small{
   \textsuperscript{*}\textbf{Correspondence:}
   Ivan Sviridov, \href{mailto:wchhiaarid@gmail.com}{\nolinkurl{wchhiaarid@gmail.com}};
   Aleksandr Nesterov, \href{mailto:LAST_AUTHOR_EMAIL}{\nolinkurl{nesterov@airi.net}}
 }
}

\begin{document}
\maketitle

\begin{abstract}
Adapting large language models (LLMs) to clinical workflows often requires costly fine-tuning or manual prompt and pipeline engineering. We study LLM-guided MAP-Elites evolution as an inference-time alternative for discovering medical decision strategies and provide an \href{https://github.com/univanxx/llm_guided_evo_medical}{implementation repository}. We formulate urgency triage, interactive consultation, and medical image classification as evolutionary searches over executable artifacts optimized by task-specific fitness functions.

Across all three settings, evolution improves over manually designed baselines under practical constraints. In triage, evolved programs increase Semigran accuracy from $77.3\%$ to $87.1\%$ and emergency recall from $0.60$ to $0.97$, while improving safety-weighted held-out MIMIC-ESI performance. In interactive consultation, evolved policies improve the accuracy--cost frontier across Llama-3, Qwen-3.5, and Gemma-4 and transfer to held-out iCRAFTMD. In PneumoniaMNIST, prompt-only evolution improves frozen MedGemma VLMs while preserving strict JSON outputs. Qualitative analysis shows that the gains come from interpretable program-level mechanisms, calibrated triage boundaries, targeted evidence acquisition, selective commitment, and finding-oriented visual decision rules, rather than superficial prompt rewording alone.
\end{abstract}

\section{Introduction}
Recent advances in deep learning, particularly large language models (LLMs), have enabled increasingly capable medical decision pipelines that combine model inference with prompting strategies, evidence acquisition, and task-specific decision rules. Such pipelines have demonstrated strong performance across clinical settings such as medical triage, image classification, and interactive clinical reasoning~\cite{masanneck2024triage,fang2025large,li2024mediq}, motivating growing interest in using LLMs to assist clinical decision-making and support healthcare workflows.

However, adapting LLMs to specialized domains such as healthcare remains computationally and operationally expensive. Fine-tuning LLMs requires substantial GPU resources, large-scale annotated datasets, and careful hyperparameter optimization~\cite{singh2025survey}. In parallel, developing effective prompting strategies and inference pipelines often depends on iterative human-in-the-loop refinement, expert intuition, and repeated empirical testing~\cite{yao2018automl,zhang2025exploring}. This manual process of refining prompts, parameters, and heuristics is time-consuming, difficult to reproduce, and often inconsistent across practitioners and application settings~\cite{beam2020challenges,ferreira2025pre}.

Recently, LLM-guided \textit{MAP-Elites} optimization has emerged as an inference-time alternative to fine-tuning and manual pipeline engineering~\cite{lehman2022evolution}. MAP-Elites is a quality-diversity algorithm that preserves high-performing candidates across behavioral dimensions rather than optimizing only a single best solution~\cite{mouret2015illuminatingsearchspacesmapping}.In this setting, a frozen LLM mutates candidate artifacts such as programs, configurations, or prompts, while task-specific evaluators score them and update the archive~\cite{lim2024largelanguagemodelsincontext}. This combination of LLM-based rewriting and archive-based selection has improved solutions across algorithmic and optimization tasks~\cite{novikov2025alphaevolvecodingagentscientific,khrulkov2025gigaevo}.


However, compared with other application domains, the use of MAP-Elites-based, LLM-driven evolutionary frameworks in medicine remains limited. Existing MAP-Elites prompt-optimization studies mainly target general-domain benchmarks~\cite{santos2025diversepromptsilluminatingprompt}, leaving open whether such methods can optimize medical decision pipelines across a broader clinical decision spectrum. We therefore study three complementary settings that span distinct modes of clinical reasoning: \textit{triage classification} for risk-sensitive urgency assessment~\cite{da2025ai}, \textit{interactive consultation} for sequential evidence acquisition under incomplete information~\cite{li2024mediq}, and \textit{medical image classification} for visual diagnosis~\cite{schouten2025navigating}.

Across these settings, we evaluate whether LLM-guided MAP-Elites evolution can improve medical decision pipelines over manually designed baselines while preserving practical constraints such as interaction cost, safety-relevant behavior, and output validity. We use task-specific candidate representations and MAP-Elites configurations, ranging from executable decision programs to prompt-producing modules to support search across heterogeneous clinical decision settings rather than a single fixed pipeline.

\paragraph{Contributions.}
Our contributions are threefold. 
\textbf{(i)} We show that LLM-guided MAP-Elites evolution can generate and refine medical decision strategies that match or outperform manually designed baselines under task-specific clinical objectives across three decision settings. 
\textbf{(ii)} We evaluate when and how these gains hold through quantitative and qualitative analyses of accuracy, safety-relevant behavior, interaction cost, cross-model transfer, held-out generalization, and ablations. 
\textbf{(iii)} We show that the gains arise from interpretable changes in decision logic, including risk-sensitive classification, targeted evidence acquisition, selective commitment, and finding-oriented visual prompts with strict structured outputs.

\section{Related Work}
\label{sec:related_work}

Evolutionary algorithms (EAs) are population-based optimizers that iteratively generate, evaluate, and select candidate solutions under a task-specific fitness function~\citep{vcrepinvsek2013exploration}. A particularly relevant family is \textit{MAP-Elites}, a quality-diversity algorithm that preserves high-performing candidates in an archive indexed by behavioral descriptors~\citep{mouret2015illuminatingsearchspacesmapping}. Rather than returning only a single best solution, MAP-Elites maintains diverse solution types while optimizing fitness within each archive cell, making it well-suited to automated pipeline design where structurally different programs or prompts may solve the same task through different strategies~\citep{lim2024largelanguagemodelsincontext}. Recent work combines this archive-based search with LLM-generated mutations: pre-trained LLMs have been used as mutation operators for discovering robot-control programs~\citep{lehman2022evolution}, as program-search engines for mathematical discovery~\citep{romera2024mathematical}, and as coding agents for larger codebases and scientific or engineering tasks~\citep{novikov2025alphaevolvecodingagentscientific}. Related work on co-evolving rationales with executable code and on open-source evolutionary optimization frameworks further suggests that LLM-guided evolutionary search is reproducible and broadly applicable~\citep{liu2024evolving,khrulkov2025gigaevo}. These studies establish LLM-guided evolution as a general program-optimization paradigm, but they do not directly address safety-sensitive medical decision pipelines.

Automated LLM optimization has begun to appear in medical machine learning. However, existing work mostly targets isolated prompts or general medical QA benchmarks rather than executable decision procedures whose prompts, control logic, interaction policy, and output constraints can all change during search~\citep{chen2025empowerevolutionarymedicalprompt,wu2025automedprompt}. This distinction is important across the three settings we study. In triage, prior work evaluates LLMs and symptom-checking systems under fixed prompts or manually designed decision procedures, rather than evolving the decision pipeline itself~\citep{williams2024llmtriage,masanneck2024triage,medask2025triage}. In medical image analysis, recent studies typically evaluate fixed radiology-inspired prompts, visual finding checklists, or manually assembled description-then-classification pipelines, rather than treating the prompt or inference procedure as an object of systematic evolutionary search~\citep{qin2022medical,pmlr-v315-byun26a}. In multi-turn clinical dialogue, question-asking and abstention strategies are likewise manually designed and implemented as fixed policy families~\citep{li2024mediq}. In contrast, we evaluate LLM-guided MAP-Elites as a unified approach for optimizing medical decision pipelines across triage classification, interactive consultation, and medical image classification, with task-specific fitness functions reflecting safety-sensitive recall, interaction cost, and structured-output validity.

\section{Clinical Tasks}
\label{sec:clinical_tasks}

\subsection{Shared Evolutionary Setup}
\label{sec:shared_evolutionary_setup}

We use GigaEvo, an open-source framework for LLM-guided program evolution with MAP-Elites search, asynchronous evaluation, rewrite-based mutation, and lineage tracking~\citep{khrulkov2025gigaevo}. Across tasks, candidate solutions are executable artifacts--decision programs, consultation policies, or prompt-producing modules--that are scored by task-specific fitness functions, selected via MAP-Elites, and rewritten by an LLM mutator. This formulation enables optimization of heterogeneous medical decision pipelines without fine-tuning the underlying task models.

The mutation step requires reliable code rewriting, structured use of fitness feedback, and medically informed reasoning. We therefore use \texttt{gpt-oss-120b} as the evolver: an open-weight sparse-MoE model with strong reported coding, reasoning, tool-use, and structured-output capabilities~\citep{agarwal2025gpt}. Its medical knowledge distillation further supports this setting, where candidate programs may elicit clinical reasoning without relying only on explicit domain knowledge bases~\citep{zhou2025enhancing}. For each task, we select final candidates based on training fitness and repeatedly re-evaluate on held-out test data to account for LLM stochasticity. Task-specific candidates and fitness functions are described below; full seed pools, optimization details, and MAP-Elites hyperparameters are in Appx.~\ref{app:clinical_tasks_details}.

\subsection{Triage}
Triage evaluates whether evolution can improve the assignment of urgency under asymmetric clinical risk. We consider two complementary formulations. The first is the Semigran vignette benchmark \cite{Semigranh3480}, a three-class symptom-checker triage task with labels \texttt{em} (emergency), \texttt{ne} (non-emergency requiring clinical evaluation), and \texttt{sc} (self-care). Because Semigran contains only 45 vignettes, we additionally evaluate transfer to the independently constructed Levine et al. vignette set \cite{Levine2023.01.30.23285067}. The second formulation is MIMIC-IV-ED-derived Emergency Severity Index (ESI) prediction \cite{gaber2025llmworkflows}, where each program receives a patient description containing history of present illness, demographics, chief complaint, and initial vitals, and predicts an ESI level from 1 to 5. This setting supports a cleaner train/audit/test protocol: programs evolve on the MIMIC-derived training pool, are selected on a fixed, balanced audit split, and are finally evaluated on a held-out test split.

Candidate triage systems are standalone Python modules that call the underlying LLM through structured helper APIs and return a prediction for each case. Evolution may therefore modify not only the prompt, but also decision decomposition, voting, retrieval, resource scoring, and rule-based post-processing. In the Semigran setting, initial programs include direct prompting, structured JSON output, chain-of-thought prompting, debate-style prompting, knowledge-base retrieval, severity scoring, and decision-tree logic. In the MIMIC-ESI setting, initial programs include article-style clinical prompts, structured ESI assessment, resource-count reasoning, self-consistency voting, and compact local ESI-reference retrieval. Detailed seed descriptions are in Appx.~\ref{app:triage_details}.

The Semigran fitness function combines overall accuracy with emergency recall ($R_{em}$), reflecting the higher clinical cost of undertriage; $R_{ne}$ and $R_{sc}$ denote non-emergency and self-care recall. For MIMIC-ESI, we optimize a safety-weighted objective combining exact ESI accuracy, range accuracy, and ESI-1 recall; range accuracy treats exact predictions and one-level safer overtriage as correct. During the final evaluation, we additionally report undertriage, severe undertriage, overtriage, and class-specific recall to expose safety-specific trade-offs. Full formulas and audit/test selection details are given in Appx.~\ref{app:triage_details}.

\subsection{Interactive Consultation}
Interactive consultation evaluates sequential evidence acquisition under incomplete information. We use the MEDIQ formulation \cite{li2024mediq}, where an Expert model receives an initially incomplete patient case, asks follow-up questions to a Patient system, and decides when the collected evidence is sufficient to answer a multiple-choice medical question. We use the iMEDQA training split \cite{app11146421} and reserve iCRAFTMD as the held-out evaluation set \cite{johri2025evaluation}. This task differs from static medical QA because the candidate must jointly learn when to ask, what to ask, and when to stop.

Each candidate program implements an abstention policy controlling whether the Expert should continue information gathering or produce a final answer. We initialize from MEDIQ-style strategies, including implicit abstention, binary criteria, and numerical or scale-based confidence estimation, optionally combined with rationale generation and self-consistency. We evaluate both direct comparability with Llama-3-8B-Instruct and Llama-3-70B-Instruct \cite{grattafiori2024llama3herdmodels} and transfer to stronger open-weight models, Qwen-3.5-27B \cite{qwen35blog} and Gemma-4-31B. Since interaction cost is central to this setting, we report answer accuracy along with average Expert-token usage; for modern-model comparisons, we also use a Borda-style \cite{lippman_voting} aggregate rank that combines accuracy and token cost. The full partial-evaluation scheme, including structured batches, lineage blending, and the follow-up-question cap, is described in Appx.~\ref{app:consultation_details}.

\subsection{PneumoniaMNIST}
PneumoniaMNIST evaluates prompt-only evolution for fixed medical vision-language models. The task is binary pediatric chest-X-ray classification from the MedMNIST collection, with labels \texttt{normal} and \texttt{pneumonia} \cite{yang2023medmnist}. We run separate experiments for the MedMNIST+ image resolutions \(28\times28\), \(64\times64\), \(128\times128\), and \(224\times224\) \cite{doerrich2025medmnistplus}, and evaluate two MedGemma multimodal models: MedGemma-4B and MedGemma-27B multimodal \cite{sellergren2025medgemma}. For each model-resolution pair, the underlying VLM, image preprocessing, label set, and JSON output parser are fixed; evolution can improve performance only by changing the prompt returned by the candidate program.

Each candidate is a Python module that returns a prompt for the fixed MedGemma classifier, which must output strict JSON with \texttt{class\_id} and \texttt{class\_name}. We initialize baselines from radiology-style prompts, select the best training seed as \textsc{Base}, evolve on a balanced 100-image training subset, and evaluate the selected Base and evolved prompts on the official test split. Full prompt seeds and optimization details are in Appx.~\ref{app:pneumoniamnist_details}.

\section{Quantitative Results}

\subsection{Triage Task Results}

\paragraph{Semigran Vignette Benchmark.}
Table~\ref{tab:evolved} shows that MAP-Elites substantially improves over the best initial Semigran program. The strongest evolved program, \texttt{SG-c1189}, reaches $87.1\%\pm2.0$ accuracy, improving over Base by $+9.8$ pp ($p<0.001$, Wilcoxon signed-rank, $K{=}10$), while increasing emergency recall from $R_{em}=0.60$ to $0.97$ and preserving useful non-emergency and self-care recall. This result addresses the main baseline failure mode: the best initial program recognizes non-emergency and self-care cases well, but misses many emergency cases. Appx.~\ref{app:cross-model-transfer} further shows that this program transfers without code changes to unseen LLM backbones.

Table~\ref{tab:comparison} contextualizes this result against published Semigran triage systems. \texttt{SG-c1189} outperforms all compared LLM and symptom-checker baselines except human physicians, while maintaining both high emergency recall ($R_{em}>0.97$) and self-care recall ($R_{sc}>0.80$). In contrast, high-recall external LLM baselines often sacrifice self-care recognition.

On the independently constructed Levine vignette set (Appx.~\ref{app:held-out-evaluation}), \texttt{SG-c1189} preserves perfect emergency recall ($R_{em}=1.00$) but drops to $75.1\%\pm2.9$ accuracy due to conservative overtriage of non-emergency cases. This difference suggests safety-oriented transfer with Semigran-specific specialization.

\begin{table}[t]
\centering
\resizebox{\columnwidth}{!}{%
\begin{tabular}{lcccccc}
\hline
\textbf{Program} & \textbf{Acc (\%)} & $\mathbf{R_{em}}$ & $\mathbf{R_{ne}}$ & $\mathbf{R_{sc}}$ & \textbf{Fit.} & $\boldsymbol{\Delta}$\textbf{Acc} \\
\hline
Base (best init.) & $77.3 {\pm} 1.8$ & 0.60 & 0.88 & 0.84 & 0.747 & --- \\
\hline
\texttt{SG-c1189} & $\mathbf{87.1 {\pm} 2.0}$ & \textbf{0.97} & 0.83 & 0.81 & \textbf{0.886} & $+9.8$ \\
\texttt{SG-a4a9e} & $86.4 {\pm} 1.9$ & 0.97 & 0.81 & 0.81 & 0.881 & $+9.2$ \\
\texttt{SG-3eed9} & $84.0 {\pm} 2.0$ & 0.93 & 0.81 & 0.78 & 0.853 & $+6.7$ \\
\texttt{SG-d49e8} & $83.3 {\pm} 2.6$ & 0.96 & 0.68 & 0.86 & 0.852 & $+6.0$ \\
\texttt{SG-1043d} & $81.8 {\pm} 1.8$ & 0.97 & 0.69 & 0.79 & 0.841 & $+4.4$ \\
\hline
\end{tabular}}
\caption{Top-5 evolved programs vs.\ best initial program on Semigran. All evolved programs significantly outperform Base ($p < 0.001$, Wilcoxon signed-rank, $K{=}10$ paired runs).}
\label{tab:evolved}
\end{table}


\begin{table}[t]
\centering
\resizebox{\columnwidth}{!}{%
\begin{tabular}{llcccc}
\hline
\textbf{System} & \textbf{Model} & \textbf{Acc (\%)} & $\mathbf{R_{em}}$ & $\mathbf{R_{ne}}$ & $\mathbf{R_{sc}}$ \\
\hline
Physicians$^a$               & Human        & 91.0             & ---  & ---  & --- \\
MedAsk$^b$                   & Propr.       & $82.7 {\pm} 4.0$   & 0.96 & 0.80 & 0.72 \\
\hline
o4-mini$^c$                  & o4-mini      & 80.4             & 0.89 & 0.81 & 0.71 \\
o3$^c$                       & o3           & 75.6             & 0.91 & 0.83 & 0.53 \\
o1$^b$                       & o1           & $73.3 {\pm} 3.5$   & 0.87 & 0.89 & 0.44 \\
GPT-4o$^b$                   & GPT-4o       & $69.3 {\pm} 1.0$   & 0.80 & 0.88 & 0.40 \\
o3-mini$^b$                  & o3-mini      & $69.3 {\pm} 2.9$   & 0.80 & 0.95 & 0.33 \\
GPT-4.5$^c$                  & GPT-4.5      & 68.9             & 0.93 & 0.83 & 0.31 \\
\hline
Base (best init.)            & gpt-oss-120b & $77.3 {\pm} 1.8$   & 0.60 & 0.88 & 0.84 \\
\textbf{\texttt{SG-c1189}}   & gpt-oss-120b & $\mathbf{87.1 {\pm} 2.0}$ & \textbf{0.97} & 0.83 & \textbf{0.81} \\
\hline
\end{tabular}}
\caption{Comparison with published triage systems on Semigran.
$^a$\,\cite{Semigranh3480};
$^b$\,MedAsk, Mar 2025 (5 runs);
$^c$\,MedAsk, Jul 2025 (single run).}
\label{tab:comparison}
\end{table}

\paragraph{MIMIC-ESI Benchmark.}

Table~\ref{tab:mimic_esi_main} reports results on the held-out MIMIC-ESI test split ($n{=}2{,}000$). Unlike Semigran, this experiment uses a train/audit/test protocol: programs evolve on stratified batches from the MIMIC training pool, are selected on a fixed audit split, and are evaluated on the test split only after selection. Figure~\ref{fig:mimic_esi_fitness} in Appx.~\ref{app:mimic-eci-evolution-dynamics} shows sustained improvement in archive fitness across generations.


The best-evolved program, \texttt{MIMIC-023}, outperforms all hand-designed baselines. It reaches $62.0\%\pm0.4$ exact accuracy, compared with $56.7\%\pm1.1$ for the strongest exact-accuracy baseline, \textsc{KB-Full-Context}, and achieves fitness $0.648\pm0.003$, compared with $0.558\pm0.031$ for the strongest baseline by fitness, \textsc{Self-Consistency-Vote}. It also improves range accuracy from $70.3\%$ to $77.0\%$ and reduces severe undertriage from $3.6\%$ to $1.2\%$ relative to \textsc{Self-Consistency-Vote}. Other evolved programs exhibit different operating points: \texttt{MIMIC-018} is close in terms of exact accuracy and fitness. In contrast, \texttt{MIMIC-011} favors a more safety-oriented profile with higher range accuracy and ESI-1 recall but lower exact accuracy.

\begin{table*}[t]
\centering
\resizebox{\textwidth}{!}{%
\begin{tabular}{lcccccc}
\hline
\textbf{Program} & \textbf{Exact (\%) $\uparrow$} & \textbf{Range (\%) $\uparrow$} & \textbf{Under. (\%) $\downarrow$} & \textbf{Severe under. (\%) $\downarrow$} & $\mathbf{R_{\mathrm{ESI}=1}}$ \textbf{$\uparrow$} & \textbf{Fitness $\uparrow$} \\
\hline
Article prompt clinical & $51.4 \pm 0.5$ & $64.7 \pm 0.1$ & $35.1 \pm 0.1$ & $5.0 \pm 0.1$ & $27.3 \pm 1.2$ & $0.523 \pm 0.004$ \\
KB-Full-Context         & $56.7 \pm 1.1$ & $65.0 \pm 1.7$ & $34.9 \pm 1.6$ & $3.7 \pm 0.4$ & $16.2 \pm 0.4$ & $0.547 \pm 0.010$ \\
Self-Consistency-Vote   & $54.1 \pm 2.5$ & $70.3 \pm 4.7$ & $29.2 \pm 4.7$ & $3.6 \pm 1.6$ & $30.4 \pm 3.5$ & $0.558 \pm 0.031$ \\
\hline
\texttt{MIMIC-011} & $53.5 \pm 2.0$ & $\mathbf{78.4 \pm 2.6}$ & $\mathbf{15.8 \pm 2.7}$ & $2.7 \pm 3.7$ & $\mathbf{79.8 \pm 0.4}$ & $0.624 \pm 0.019$ \\
\texttt{MIMIC-018} & $60.9 \pm 0.8$ & $75.4 \pm 0.5$ & $22.8 \pm 0.6$ & $\mathbf{1.0 \pm 0.0}$ & $56.0 \pm 1.3$ & $0.640 \pm 0.008$ \\
\texttt{MIMIC-023} & $\mathbf{62.0 \pm 0.4}$ & $77.0 \pm 0.4$ & $21.6 \pm 0.5$ & $1.2 \pm 0.1$ & $52.9 \pm 1.0$ & $\mathbf{0.648 \pm 0.003}$ \\
\hline
\end{tabular}}
\caption{MIMIC-ESI held-out results on the test split. Range metric includes exact predictions and one-level safer overtriage; Under.\ denotes undertriage.}
\label{tab:mimic_esi_main}
\end{table*}

Table~\ref{tab:mimic_esi_gaber} in Appx.~\ref{app:contextual-comparison-on_mimic} contextualizes \texttt{MIMIC-023} against the MIMIC-derived ESI results of Gaber et al.~\citep{gaber2025llmworkflows}. This comparison is not a direct replication, since the systems use different base models, retrieval setups, and evaluation splits. Still, \texttt{MIMIC-023} is competitive with Claude~3 Sonnet on exact accuracy ($62.0\%$ vs.\ $61.65\%$), improves over it on range accuracy ($77.0\%$ vs.\ $74.55\%$), and approaches the RAG-assisted range accuracy reported in the reference study.

\subsection{Interactive Consultation Task Results}



\begin{table}[t]
\centering
\footnotesize
\setlength{\tabcolsep}{2.2pt}
\begin{tabular*}{\columnwidth}{@{\extracolsep{\fill}}lcccc@{}}
\hline
\textbf{Policy} &
\textbf{Acc. (\%) $\uparrow$} &
\textbf{Avg. tok. $\downarrow$} &
\textbf{$\Delta$Acc.} &
\textbf{Tok. red.} \\
 & & & \textbf{(pp)} & \textbf{(\%)} \\
\hline
\multicolumn{5}{@{}l}{\textbf{Llama-3-8B}} \\
Num. cutoff
& $45.1 \pm 0.6$
& $2788 \pm 7$
& -- & -- \\
Best from \mediq{}
& $45.8 \pm 1.4$
& --
& $+0.7$ & -- \\
Evolved
& $\mathbf{48.2} \pm 0.4$
& $\mathbf{289} \pm 2$
& $+3.1$
& $89.6$ \\
\hline
\multicolumn{5}{@{}l}{\textbf{Llama-3-70B}} \\
Binary
& $58.6 \pm 0.3$
& $1585 \pm 1$
& -- & -- \\
Best from \mediq{}
& $60.9 \pm 1.4$
& --
& $+2.3$ & -- \\
Evolved
& $\mathbf{62.2} \pm 0.3$
& $\mathbf{514} \pm 1$
& $+3.6$
& $67.6$ \\
\hline
\multicolumn{5}{@{}l}{\textbf{Qwen-3.5-27B}} \\
Num. cutoff
& $71.1 \pm 1.1$
& $2100 \pm 10$
& -- & -- \\
Evolved
& $\mathbf{73.6} \pm 0.6$
& $\mathbf{961} \pm 8$
& $+2.5$
& $54.2$ \\
\hline
\multicolumn{5}{@{}l}{\textbf{Gemma-4-31B}} \\
Num. cutoff
& $\mathbf{76.4} \pm 0.4$
& $1412 \pm 1$
& -- & -- \\
Evolved
& $75.6 \pm 0.5$
& $\mathbf{104} \pm 3$
& $-0.8$
& $92.6$ \\
\hline
\end{tabular*}
\caption{\imedqa{} test accuracy--cost summary on the \mediq{} benchmark. For each model, evolved is compared against the strongest non-evolved baseline with reported token usage; $\Delta$Acc.\ and Tok.\ red.\ are computed relative to that baseline. Full per-baseline results are reported in Appx.~\ref{app:mediq_full_imedqa}.}
\label{tab:mediq_imedqa_summary}
\end{table}


\paragraph{\imedqa{} Results.}
Table~\ref{tab:mediq_imedqa_summary} summarizes the main \imedqa{} accuracy--cost comparison across expert models; full per-baseline results are provided in Appx.~\ref{app:mediq_full_imedqa}. On Llama-3-8B and Llama-3-70B, the evolved strategy improves over the strongest non-evolved comparators with reported token usage by $+3.1$ and $+3.6$ percentage points, respectively, while reducing average Expert-token usage by $89.6\%$ and $67.6\%$. It also exceeds the best accuracies reported by the original \mediq{} configurations, improving from $45.8\%$ to $48.2\%$ on Llama-3-8B and from $60.9\%$ to $62.2\%$ on Llama-3-70B. However, the use of tokens for those reference configurations is not stated in the original paper.

The same pattern largely transfers to stronger open-weight models. On Qwen-3.5-27B, the evolved strategy achieves the best accuracy, improving over the numerical-cutoff baseline from $71.1\%$ to $73.6\%$ while reducing token usage from $2100$ to $961$ tokens. On Gemma-4-31B, the evolved strategy trades a small accuracy drop relative to the numerical-cutoff baseline ($75.6\%$ vs.\ $76.4\%$) for a much larger reduction in token usage ($104$ vs.\ $1412$ tokens, $92.6\%$ fewer). Thus, evolution does not simply increase dialogue length: it finds policies that preserve or improve decision quality while moving the strategy toward a better accuracy--cost frontier.

\paragraph{Held-out Evaluation on \icraftmd.}
To assess whether the evolved strategy generalizes beyond \imedqa{} used during evolution, we evaluated the same mutants on the held-out \icraftmd{} split. Tables~\ref{tab:icraftmd_llama_metrics} and~\ref{tab:icraftmd_modern_metrics} in Appx.~\ref{app:interactive-consultation-task-results} show that the evolved strategy preserves its accuracy--cost advantage across both model families, consistently achieving the highest aggregate rank of 7 under the Borda-style score. This observation again indicates that the strategy does not improve accuracy by simply increasing the number of interaction tokens, but instead maintains strong decision quality while keeping interaction cost low.


\paragraph{Ablation Study.}
Table~\ref{tab:mediq_ablation} in Appx.~\ref{app:inter-cons-ablation-study} shows that both fitness-stabilization components matter: removing structured batch composition reduces Llama-3-8B accuracy from $48.2\%$ to $44.8\%$, and further removing lineage blending lowers it to $43.0\%$, indicating that both components help stabilize candidate ranking under partial evaluation.


\subsection{Results for Classification of PneumoniaMNIST}

Table~\ref{tab:medgemma_pneumonia_resolution_metrics} compares the best initial prompt and the best evolved prompt on the official PneumoniaMNIST test split across all image resolutions. For each model--resolution pair, \textit{Base} denotes the highest-fitness initial seed prompt and \textit{Final} denotes the highest-fitness evolved prompt selected after evolution.

\begin{table*}[t]
\centering
\small
\begin{tabular}{llcccc}
\hline
\textbf{Program} & \textbf{Resolution} & \multicolumn{2}{c}{\textbf{MedGemma-27B}} & \multicolumn{2}{c}{\textbf{MedGemma-4B}} \\
\cline{3-4} \cline{5-6}
 & & \textbf{Acc. (\%) $\uparrow$} & \textbf{Macro-F1 (\%) $\uparrow$} & \textbf{Acc. (\%) $\uparrow$} & \textbf{Macro-F1 (\%) $\uparrow$} \\
\hline
Base  & $28 \times 28$   & $52.53 \pm 0.24$ & $52.18 \pm 0.23$ & $37.50 \pm 0.00$ & $27.27 \pm 0.00$ \\
Final & $28 \times 28$   & $63.40 \pm 0.30$ & $63.25 \pm 0.29$ & $68.01 \pm 0.58$ & $67.35 \pm 0.56$ \\
\hline
Base  & $64 \times 64$   & $52.92 \pm 0.06$ & $51.10 \pm 0.05$ & $38.62 \pm 0.00$ & $29.45 \pm 0.00$ \\
Final & $64 \times 64$   & $65.80 \pm 0.08$ & $65.76 \pm 0.08$ & $63.24 \pm 0.31$ & $53.18 \pm 0.36$ \\
\hline
Base  & $128 \times 128$ & $75.06 \pm 0.08$ & $75.03 \pm 0.08$ & $50.32 \pm 0.00$ & $47.10 \pm 0.00$ \\
Final & $128 \times 128$ & $83.81 \pm 0.10$ & $81.68 \pm 0.13$ & $62.12 \pm 0.35$ & $63.91 \pm 0.33$ \\
\hline
Base  & $224 \times 224$ & $83.33 \pm 0.00$ & $82.96 \pm 0.00$ & $63.04 \pm 0.08$ & $62.51 \pm 0.09$ \\
Final & $224 \times 224$ & $84.46 \pm 0.14$ & $83.88 \pm 0.15$ & $72.50 \pm 0.37$ & $72.50 \pm 0.38$ \\
\hline
\end{tabular}
\caption{PneumoniaMNIST test results across image resolutions (5 runs). Base and Final denote the best initial and evolved prompts selected by training fitness, respectively.}
\label{tab:medgemma_pneumonia_resolution_metrics}
\end{table*}


Across resolutions, the evolved prompt improves test accuracy and macro-F1 for both MedGemma model families. Resolution has a strong effect, even though all images are resized to the MedGemma input size before inference: higher source resolution preserves more diagnostic structure before resizing. For MedGemma-27B, both Base and Final improve sharply from $64 \times 64$ to $128 \times 128$, after which performance largely saturates; the best result is obtained at $224 \times 224$, with $84.46\%$ accuracy and $83.88\%$ macro-F1. The evolutionary gain is largest when the Base prompt is still weak, and becomes smaller at $224 \times 224$, where the Base prompt already reaches $83.33\%$ accuracy.

MedGemma-4B is more resolution-sensitive in absolute performance, but also benefits more from prompt evolution at low and medium resolutions. Its Base prompt remains below $51\%$ accuracy up to $128 \times 128$, whereas Final reaches $68.01\%$ at $28 \times 28$ and $63.24\%$ at $64 \times 64$. At $224 \times 224$, Final reaches the strongest 4B result, $72.50\%$ accuracy, and $72.50\%$ macro-F1. This result suggests that, for the smaller VLM, the prompt plays a larger role in setting the decision threshold between normal and pneumonia, especially when low-resolution images make visual findings subtle or ambiguous.

\paragraph{Comparison with prior PneumoniaMNIST results.}
Compared with prior PneumoniaMNIST work, our results should be interpreted as a data-efficient prompt-only adaptation result rather than as a substitute for supervised fine-tuning or multi-model diagnostic pipelines. A recent pre-consultation dialogue framework evaluates several VLM-based settings on PneumoniaMNIST: plain MedGemma-4B zero-shot prompting achieves $45.8\%$ accuracy / $40.8\%$ F1, and chain-of-thought prompting achieves $46.3\%$ /
$41.5\%$; the corresponding dialogue-based zero-shot variant reaches $87.8\%$ /
$87.6\%$ with MedGemma-4B and $94.6\%$ / $94.3\%$ with Qwen-2.5-VL \citep{lokesh2026patientvlm}. Thus, our evolved single-prompt MedGemma setup substantially outperforms the plain MedGemma-4B zero-shot and chain-of-thought baselines reported in that study, while remaining below its manually engineered multi-agent dialogue pipeline.

Task-adapted medical VLMs and specialist classifiers provide a complementary upper-reference regime. A diagnosis-guided, bootstrapped medical VLM reported to achieve $87.0\%$ accuracy/$87.3\%$ macro-F1, and its retrieval-augmented variant $ 92.9\%$/$92.6\%$ \citep{he2024gsco}. Task-specific visual instruction tuning reaches $93.1\%$ accuracy / $92.3\%$ macro-F1, while a fully supervised
task-specific ViT classifier reaches $96.8\%$ / $96.5\%$
\citep{bai2024vitask}. These results illustrate the expected gap between prompt-only adaptation of a frozen VLM and methods that use task-specific
training, retrieval, specialist components, or multi-turn inference machinery. In contrast, our method keeps MedGemma frozen and evolves only the prompt, using a fixed $ 100$-image-labeled subset for fitness evaluation.

\section{Qualitative Results}
\label{sec:qualitative_results}

The quantitative results show that LLM-guided evolution improves accuracy, safety-sensitive behavior, interaction cost, and output validity across the three tasks. We next analyze the best evolved programs to understand \emph{how} these improvements arise. This analysis is important because the search space contains executable decision artifacts rather than only natural-language prompts: evolution can change not only wording, but also voting rules, stopping criteria, retrieval use, fallback behavior, and deterministic post-processing. Across tasks, we find that the evolved candidates are interpretable and that their gains come from program-level mechanisms: calibrated triage boundaries, targeted evidence acquisition, selective commitment, and finding-oriented visual decision rules. Extended qualitative analyses and representative code excerpts are provided in Appendices~\ref{app:extended_qualitative_analysis} and~\ref{app:qualitative_snippets}.

\subsection{Triage}

In triage, evolution primarily shifts the decision system's operating point. On Semigran, the strongest initial program, \textsc{Base}, already recognizes non-emergency and self-care cases reasonably well, but misses many emergencies ($R_{\mathrm{em}}{=}0.60$). The best evolved program, \texttt{SG-c1189}, shifts this boundary toward safety, reaching $R_{\mathrm{em}}{=}0.97$ while preserving useful non-emergency and self-care recall ($R_{\mathrm{ne}}{=}0.83$, $R_{\mathrm{sc}}{=}0.81$). This observation is not the degenerate solution of predicting emergencies for most cases: compared with high-recall external systems, the evolved program retains substantially stronger self-care recognition. Qualitatively, the change is a recalibration of the triage boundary rather than a formatting improvement.

A second Semigran observation is that the best evolved program does not rely on the explicit triage knowledge base. Although several seed programs inject ESI, CTAS, or START reference material, the winning candidate is a compact prompt-based wrapper. In this setting, evolution improves decision framing and class-boundary calibration more than it benefits from additional retrieved context. However, the Levine transfer evaluation also shows the cost of this specialization: \texttt{SG-c1189} preserves perfect emergency recall on independently constructed vignettes, but overtriages many non-emergency cases. Thus, the Semigran result is best interpreted as successful safety-oriented specialization, not as evidence that a single evolved vignette wrapper solves triage robustness.

The MIMIC-ESI setting reveals a complementary pattern. The best held-out program, \texttt{MIMIC-023}, combines local ESI reference retrieval, structured voting, and resource-sensitive reasoning, yielding the best exact accuracy and overall fitness on the MIMIC test split. Other evolved candidates expose different clinically meaningful points on the search landscape. For example, \texttt{MIMIC-011} is more conservative: it improves safety-oriented behavior on MIMIC by increasing ESI-1 recall and reducing undertriage, but at the expense of exact accuracy. Under external transfer, this same candidate collapses into an all-emergency classifier because a high-acuity rule treats missing systolic blood pressure as zero. This failure mode is informative: evolution can discover plausible safety heuristics, but such heuristics need explicit robustness constraints around missingness and distribution shift.

\subsection{Interactive Consultation}

For interactive consultation, evolution discovers policies that improve the accuracy--cost trade-off by changing when the agent commits, what it asks, and how it stabilizes final decisions. The Llama-3-8B evolved program keeps the low-cost structure of implicit abstention, but derives a confidence estimate from the answer-vote distribution and commits only under a supermajority threshold. This result preserves the efficiency of a single-pass implicit strategy while adding a reliability check before the final answer. The Llama-3-70B evolved program changes the question-selection policy: instead of choosing follow-up questions only by vote frequency or at random, it scores questions for clinically informative attributes such as onset, duration, severity, risk factors, and family history. This change shifts the consultation from generic information-gathering to high-yield evidence acquisition.

The transferred evolved programs show two further mechanisms. The Qwen-3.5-27B candidate tracks evidence at the answer-option level and asks fallback questions about the least-supported hypothesis, implementing an evidence-balancing policy across competing diagnoses or treatments. The Gemma-4-31B candidate applies self-consistency selectively at forced commitment: it samples multiple final answers only when the interaction budget is exhausted, reducing variance at the most consequential step without increasing sampling throughout the dialogue. Together, these mechanisms explain why the evolved strategies improve the accuracy--cost balance: they do not simply ask more questions, but allocate computation to uncertainty, discrimination, and final commitment.

\subsection{Classification of PneumoniaMNIST}

For PneumoniaMNIST, evolution operates only at the prompt-program level: the MedGemma model, image preprocessing, label set, and JSON parser remain fixed. The strongest initial prompts already enforce structured output, but the evolved prompts change the clinical framing of the visual task. The most consistent shift is from label-first classification to finding-oriented image assessment. Evolved prompts cast the model as a pediatric chest-radiograph reader and instruct it to inspect both lung fields for pneumonia-compatible findings such as focal or lobar air-space opacity, consolidation, patchy or perihilar infiltrates, interstitial patterns, air bronchograms, pleural effusion, and left--right asymmetry. The final prompt then maps observed findings to the required JSON label, preserving output validity while changing the internal decision procedure.

Evolution also tunes the decision threshold to the model and image resolution. Several 27B prompts favor pneumonia in the face of ambiguity, which helps improve sensitivity when subtle opacities are present. In contrast, the 4B $28{\times}28$ final prompt becomes more conservative, requiring stronger visual evidence and defaulting to normal under uncertainty. This observation suggests that the search is not merely adding medical terminology. It recombines task-level radiological hints into operational checklists and then sharpens them into model- and resolution-specific decision rules. In this sense, the qualitative effect of evolution is to turn a fixed VLM prompt into an executable visual decision policy whose threshold is adapted to the observed error profile.

\section{Conclusions}


We presented LLM-guided MAP-Elites evolution as an inference-time approach for adapting medical decision pipelines without fine-tuning the underlying task models. Across urgency triage, interactive consultation, and PneumoniaMNIST classification, the same optimization framework improved manually designed baselines while operating over different executable artifacts: triage programs, consultation policies, and prompt-producing modules for frozen vision-language models.

The results show that evolutionary search can improve clinically relevant operating points, not only aggregate accuracy. In Semigran triage, evolution increased accuracy from 77.3\% to 87.1\% and emergency recall from 0.60 to 0.97. In MIMIC-ESI, the best evolved program improved held-out exact accuracy, range accuracy, and safety-weighted fitness relative to hand-designed baselines. In interactive consultation, evolved policies improved the accuracy--cost frontier across Llama-3, Qwen-3.5, and Gemma-4, and transferred to held-out iCRAFTMD. In PneumoniaMNIST, prompt-only evolution improved frozen MedGemma models across image resolutions while preserving strict JSON output validity.

Qualitative analysis suggests that these gains come from interpretable program-level mechanisms rather than superficial prompt edits. Evolved systems recalibrate triage boundaries, combine retrieval and safety-biased voting, acquire more targeted consultation evidence, decide more selectively when to commit, and transform fixed VLM calls into finding-oriented visual decision procedures with model- and resolution-specific thresholds. This inspectability is especially important in medical settings, where candidate strategies must be auditable as decision procedures rather than treated as opaque prompts.

Evolution also introduces a non-negligible search cost. As reported in Appx.~\ref{app:evo_costs}, this cost depends on the task structure: MIMIC-ESI is dominated by multi-call candidate evaluation, interactive consultation by multi-turn textual contexts, and PneumoniaMNIST by repeated image evaluations. However, these costs are incurred during inference-time search rather than during model training, making this a practical alternative for a small-to-medium number of optimization runs when frozen models must be adapted under explicit metrics and structured-output constraints.

Overall, our findings suggest that LLM-guided evolution is a promising tool for discovering effective, inspectable, and adaptable medical decision strategies across heterogeneous clinical tasks. It is not a substitute for prospective validation, robust calibration, or safety governance. However, it provides a reproducible way to search over prompts, control logic, and decision rules that align with clinically motivated objectives.

\section*{Limitations}

\paragraph{Semigran benchmark size and overfitting.}
The Semigran triage results should be interpreted with particular care. Semigran is a compact and widely used public vignette benchmark, but it contains only 45 cases; therefore, the same set serves as both the evolutionary target and the main evaluation benchmark. This fact makes overfitting a central limitation. The Levine evaluation partially addresses this concern by testing transfer to independently constructed vignettes, where the evolved program preserves emergency recall but shifts toward conservative overtriage; however, it is not a substitute for a larger held-out benchmark from the same distribution.

\paragraph{Retrospective MIMIC-ESI benchmark.}
Our MIMIC-ESI study is based on a retrospective MIMIC-derived benchmark rather than a prospective clinical deployment. This setup enables a reproducible train/audit/test protocol, but improved offline ESI prediction does not by itself establish improved patient outcomes, workflow utility, or safety in real emergency-department use. Several limitations also arise from the benchmark construction itself: following the source-side preprocessing~\citep{gaber2025llmworkflows}, our corpus contains no observed ESI-5 cases and very few ESI-4 cases in the article-like held-out split. The training loader draws balanced batches across present classes, but cannot generate examples for absent labels. Consequently, although the task is defined over five ESI levels, our evidence is strongest for observed high- and mid-acuity cases and weak for the low-acuity tail. In addition, the benchmark uses history-of-present-illness text extracted from discharge notes, which may contain information unavailable at first-contact triage.

\paragraph{Safety-weighted objectives and inference budget.}
The optimization objective encodes task-specific value judgments. Alongside exact accuracy, we optimize range accuracy, treating exact predictions and one-level safer overtriage as correct, and explicitly rewarding ESI-1 recall. These choices reflect the asymmetric clinical cost of undertriage, but they are not neutral over all error types and may favor systems that trade specificity for safety. Moreover, candidate programs can differ in inference budget: multi-stage prompting, voting, and retrieval may require more model calls than one-shot baselines. Accuracy gains should therefore be interpreted together with latency and cost.

\paragraph{Reference-system and cross-benchmark comparisons.}
Our local ESI reference KB is useful for ablations, but it is not a recreation of the PubMed-scale RAG store used by Gaber et al.~\citep{gaber2025llmworkflows}; consequently, retrieval experiments should be interpreted as comparisons among candidate wrapper designs rather than as a direct reproduction of the reference paper's RAG system. Similarly, external evaluation of the Semigran and Levine tests focuses on transfer rather than strict task equivalence. Mapping ESI levels 1--2 to \texttt{em}, level 3 to \texttt{ne}, and levels 4--5 to \texttt{sc} enables comparison with symptom-checker benchmarks, but collapses the finer-grained ESI structure used during training. These external vignette sets can reveal robustness beyond the MIMIC source distribution, yet they do not replace prospective, multi-site validation on contemporary emergency-department data.

\paragraph{PneumoniaMNIST probability metrics.}
For PneumoniaMNIST, we report accuracy and macro-F1 but not ROC-AUC, because MedGemma is a generative VLM and does not natively provide calibrated, thresholdable class scores. Estimating ROC-AUC would require an additional scoring and calibration protocol beyond our prompt-only evaluation.

\section*{Ethics Statement}
This work uses only publicly available or access-controlled research datasets under their respective research terms, accessed only after the required credentialing and data-use approval. We do not redistribute protected health information, patient records, dataset contents, or access-controlled materials, and release only code, prompts, aggregate metrics, and derived experimental outputs that do not identify individuals.

All experiments are retrospective and benchmark-based. The proposed systems are research prototypes for medical decision-pipeline optimization, not tools for clinical deployment or autonomous medical decision-making. Given the safety-sensitive setting, we report clinically relevant trade-offs, including emergency recall, undertriage, severe undertriage, overtriage, and interaction cost where applicable.



\bibliography{custom}

\clearpage
\appendix
\twocolumn[

\vskip\intextsep
\noindent\begin{minipage}{\textwidth}
\centering
\small
\resizebox{\textwidth}{!}{\begin{tabular}{p{0.17\textwidth} p{0.40\textwidth} p{0.25\textwidth} p{0.14\textwidth}}
\hline
\textbf{Parameter} &
\textbf{Triage} &
\textbf{Interactive consultation} &
\textbf{PneumoniaMNIST} \\
\hline

MAP-Elites setup &
\textbf{Semigran:} two-island;\newline
\textbf{MIMIC-ESI:} two-island &
Single-island &
Single-island \\

Archive axes &
\textbf{Semigran:} Island~1: accuracy $\times$ $R_{em}$; Island~2: fitness $\times$ accuracy.\newline
\textbf{MIMIC-ESI:} Island~1: exact accuracy $\times$ range accuracy; Island~2: fitness $\times$ $R_{\mathrm{ESI}=1}$ &
Fitness $\times$ mean Expert-token usage &
Fitness $\times$ validity \\

Binning &
\textbf{Semigran:} Island~1: $60 \times 40$; Island~2: $50 \times 50$.\newline
\textbf{MIMIC-ESI:} Island~1: $60 \times 40$; Island~2: $50 \times 40$ &
$50 \times 100$ &
$150 \times 2$ \\

Archive cap &
75 programs per island for both triage settings &
75 programs &
75 programs \\

Archive score &
\textbf{Semigran:} Island~1: $0.85 \cdot \text{accuracy} + 0.15 \cdot R_{em}$; Island~2: $0.7 \cdot \text{fitness} + 0.3 \cdot \text{accuracy}$.\newline
\textbf{MIMIC-ESI:} Island~1: $0.7 \cdot \text{exact} + 0.3 \cdot \text{range}$; Island~2: $0.8 \cdot \text{fitness} + 0.2 \cdot R_{\mathrm{ESI}=1}$ &
Fitness score &
Fitness score \\


Migration &
Every 25 generations; 10\% migration rate; up to 5 migrants/island for both triage settings &
-- &
-- \\

Elites selected per generation &
5 for both triage settings &
5 &
5 \\

Mutants per generation &
8 for both triage settings &
4 &
8 \\


Stopping rule &
\textbf{Semigran:} 600 generations;\newline
\textbf{MIMIC-ESI:} 300 generations &
24-hour wall-clock budget &
30 generations \\

\hline
\end{tabular}}
\captionof{table}{MAP-Elites configurations across the three medical tasks.}
\label{tab:map_elites_configs}
\end{minipage}
\par\vskip\intextsep
]

\section{Clinical Tasks Details}
\label{app:clinical_tasks_details}
\subsection{Triage Task}
\label{app:triage_details}

Triage assigns patients to urgency levels that determine the required speed and intensity of medical care~\cite{robertson2006evolution}, with prior work on LLM-based triage spanning several increasingly realistic settings. Masanneck et al.~\citep{masanneck2024triage} benchmark LLMs on standardized emergency vignettes against human raters; Colakca et al.~\citep{colakca2024esi} evaluate ChatGPT on real-time emergency-department admissions under ESI principles; and Williams et al.~\citep{williams2024llmtriage} study clinical-acuity ranking from emergency-department notes rather than end-to-end ESI prediction. Gaber et al.~\citep{gaber2025llmworkflows} evaluate Claude-family models together with a RAG-assisted workflow on 2{,}000 MIMIC-IV-derived cases for ESI prediction, referral, and diagnosis. Together, these studies show that LLM-based triage has been examined across vignette benchmarks, real-world emergency-department data, note-based acuity assessment, and retrieval-augmented clinical workflows.

\subsubsection{Task Setup}
We study triage in two complementary settings. The first is the Semigran benchmark~\cite{Semigranh3480}, the most widely used public standard for evaluating symptom-checker triage accuracy. It is a three-class urgency classification task over 45 clinical vignettes evenly distributed across \texttt{em} (emergency-requires immediate care), \texttt{ne} (non-emergency-warrants clinical evaluation within days), and \texttt{sc} (self-care-manageable without professional intervention). As a held-out generalization check for this vignette setting, we additionally evaluate on the Levine et al.\ (2023)~\cite{Levine2023.01.30.23285067} vignette set ($n{=}48$), which was constructed independently of Semigran and is discussed in Section 4.1.

The second setting is a MIMIC-IV-ED-derived~\cite{PhysioNet-mimic-iv-ed-2.2} Emergency Severity Index (ESI) prediction task, following the benchmark construction of Gaber et al.~\citep{gaber2025llmworkflows} and its public code release\footnote{\url{https://github.com/BIMSBbioinfo/medLLMbenchmark}}. Here each program receives a patient description containing history of present illness, demographics, chief complaint, and initial vitals, and predicts an ESI level from 1 to 5. This task is more clinically grounded than the vignette benchmark and, crucially, supports a cleaner train/audit/test protocol. We evolve on the MIMIC-derived training pool ($n{=}25{,}330$; ESI counts $2{,}827/11{,}625/10{,}763/115$ for levels 1--4), use a fixed balanced audit split ($n{=}460$; 115 examples per observed class) for candidate selection, and reserve an article-like held-out split of 2{,}000 cases for final evaluation. The held-out split is not used for broad candidate selection.

\subsubsection{Candidates and Baselines}

Each candidate triage program is a standalone Python module that receives a list of patient descriptions and returns a structured prediction for each case. In the Semigran setting, programs output one of \texttt{em}, \texttt{ne}, or \texttt{sc}; in the MIMIC-ESI setting, they output \texttt{predicted\_acuity} in $\{1,\dots,5\}$. Programs interact with the underlying language model through an \texttt{ask\_llm()} or \texttt{ask\_llm\_json()} API; the evolutionary process searches over the \emph{program logic} that structures and refines interactions with the model, including prompt construction, decision decomposition, retrieval, voting, resource scoring, and rule-based post-processing. The base model is the open-source \texttt{gpt-oss-120b} served via an OpenAI-compatible endpoint.

For the Semigran experiment, \textsc{Base} is a hand-designed reference wrapper, while the remaining initial programs are LLM-assisted seed wrappers generated to provide a diverse starting population. These seeds cover direct prompting, structured JSON output, chain-of-thought prompting, differential diagnosis, debate-style prompting, knowledge-base retrieval, severity scoring, and simple decision-tree logic. For the MIMIC-ESI experiment, we initialize evolution from a similar mixture of manually specified reference programs and LLM-assisted seed programs. Several seeds are adapted from the prompting style and ESI evaluation setup of Gaber et al.~\citep{gaber2025llmworkflows}, including article-style clinical prompts; others implement structured ESI assessment, resource-count reasoning, self-consistency voting, and local ESI knowledge-base retrieval. The local ESI knowledge base contains compact reference material and is used to compare candidate wrapper designs; it is not a recreation of the PubMed-scale RAG system used in the reference study.

For contextual comparison we report results alongside publicly available triage benchmark numbers from the MedAsk triage benchmark~\cite{medask2025triage}, which evaluates triage performance of large language models and symptom-checking systems on the same Semigran vignettes.

\subsubsection{Optimization Objective}

The optimization objective or fitness score is a composite score:
\[
\text{fitness} = 0.85 \cdot \text{accuracy} + 0.15 \cdot \text{recall}_{em}.
\]
The asymmetric weighting is motivated by the clinical cost structure of triage errors: \emph{undertriage} (missing a genuine emergency) carries significantly higher risk than \emph{overtriage} (sending a non-urgent patient to the emergency department). The $\text{recall}_{em}$ component explicitly penalizes undertriage by rewarding programs that correctly identify life-threatening cases, even at a modest cost to overall accuracy. The 0.85/0.15 split was chosen so that a program with perfect accuracy ($\text{fitness}{=}1.0$) is preferred over one that achieves perfect emergency recall at the expense of other classes, while still providing sufficient selective pressure to drive $\text{R}_{em}$ upward. A validity gate requires accuracy $\geq 50\%$ for a program to enter the archive, which prevents degenerate solutions (e.g., always predicting \texttt{em}).

For the MIMIC-ESI experiment, we use a safety-weighted objective:
\[
\text{fitness} =
0.65 \cdot \text{exact}
+ 0.25 \cdot \text{range}
+ 0.10 \cdot R_{\mathrm{ESI}=1}.
\]
Exact accuracy requires matching the recorded ESI label. Range accuracy follows the source benchmark convention: exact predictions are correct, and one-level safer overtriage is also accepted (e.g., predicting ESI~2 for a true ESI~3 case). We additionally track undertriage, severe undertriage, overtriage, and ESI-1 recall. During evolution, each candidate is evaluated on a fresh random class-balanced sample from the full MIMIC training pool rather than on a single fixed subset. This makes individual fitness estimates noisy, but allows many generations of search to cover the full source distribution while avoiding the cost of evaluating every mutant on all 25{,}330 training cases. After evolution, top candidates are re-evaluated on the fixed balanced audit split and only then on the held-out article-like test set.


\subsection{Interactive Consultation Task}
\label{app:consultation_details}
In this paper, we study interactive consultation as a medical information-seeking task in which a model receives an initially incomplete patient case, asks follow-up questions, and decides when the collected evidence is sufficient to produce a final answer. This follows the \mediq{} benchmark formulation, which departs from static medical QA by casting clinical reasoning as an ask-or-answer process under a limited interaction budget~\cite{li2024mediq}. This framing is consistent with work on conversational diagnostic AI, which emphasizes the need to evaluate history-taking and diagnostic reasoning in realistic multi-turn consultations~\cite{tu2025towards}, while recent work on clinical question-asking highlights the importance of asking clear, relevant, and diagnostically useful follow-up questions~\cite{li2025alfa}. Together, these perspectives motivate consultation strategies that jointly determine what information to request and when the available evidence is sufficient for a reliable decision.

\subsubsection{Task Setup}
For the telemedicine consultation setting, we adopt the \mediq{} benchmark, which models interactive clinical reasoning as a multi-turn dialogue between a Patient system and an Expert system~\cite{li2024mediq}. At the start of each case, the Expert receives basic patient information $k_0$ (age, gender, and chief complaint), while the complete patient record set is multiple complaints $K=\{k_0,k_1,\dots,k_n\}$. At the start of the $t$-th turn, the Expert may ask the Patient a follow-up question $q_t$, to which the Patient system responds with a set of facts $r_t \subseteq K$ derived from the underlying record, updating the Expert’s knowledge to $K_t = K_{t-1} \cup r_t$. The objective is to iteratively acquire sufficient information to recover the subset of clinically relevant facts $K^* \subseteq K$, after which the Expert produces a final answer to a multiple-choice medical question (e.g., diagnosis or treatment decision). The goal is to learn an Expert policy that efficiently gathers informative evidence prior to making a prediction. The benchmark is constructed from \medqa{}~\cite{app11146421} and \craftmd{}~\cite{johri2025evaluation} datasets, converted into an interactive setting as \imedqa{} and \icraftmd{} using GPT-3.5~\cite{openai_gpt35turbo_docs} as the Patient model. To ensure a fair comparison, we follow the original \mediq{} setup by using the same Patient model and evaluating consultation strategies by mean answer accuracy across evaluated cases.

\subsubsection{Candidates and Baselines}
Each candidate program corresponds to a distinct abstention strategy, i.e., a policy that determines whether the Expert should produce a final answer or continue information gathering. Strategies are implemented via prompt templates and corresponding program logic that control the interaction with the underlying LLM. Specifically, we adopt as base strategies those proposed in \mediq{}, including implicit abstention decisions, binary criteria, and graded confidence estimation (numerical or Likert-scale scoring). In addition, candidates may incorporate two optional enhancements: rationale generation (explicit intermediate reasoning) and self-consistency (up to 3 times), which aggregates decisions across multiple sampled reasoning trajectories. As reported in the \mediq{} paper, the strongest-performing configuration combines scale-based abstention with rationale generation and self-consistency over three sampled reasoning trajectories.

We evaluate the candidate and baseline strategies with two groups of Expert models. First, we use Llama-3-8B-Instruct and Llama-3-70B-Instruct~\cite{grattafiori2024llama3herdmodels} to enable direct comparison with the \mediq{} baselines. Second, to assess whether the evolved strategy transfers to stronger modern open-weight models, we further evaluate it on Qwen-3.5-27B~\cite{qwen35blog} and Gemma-4-31B-it~\cite{farabet2026gemma4}, selected for their strong performance among similarly sized open-source LLMs in Arena AI\footnote{\url{https://arena.ai/leaderboard/text?license=open-source}} at the time of writing. Since all evaluated models are used in instruction-tuned variants or configurations, we omit instruction-tuning suffixes and refer to them by their base model names hereafter.

Since both accuracy and interaction cost are central to our setting, we compare methods using answer accuracy and average Expert-token usage. For the modern-model comparison, we additionally use a Borda-style ranking~\cite{lippman_voting} that scores methods separately by accuracy and average token usage, assigns positional scores from 4 to 0, and sums them across the two metrics, treating both objectives as equally important while remaining robust to their different scales.

\subsubsection{Optimization Objective}
We define fitness score as mean answer accuracy across evaluated cases, following the evaluation protocol used in the \mediq{} paper, and perform evolution on the \imedqa{} training split, while \icraftmd{} (140 cases) is reserved as a held-out evaluation set due to its smaller size. However, evaluating each candidate requires multi-turn interaction with an LLM across all 1,273 training cases, which is computationally expensive. To address this, our evolutionary configuration is designed to (i) keep evaluation tractable and (ii) ensure reliable candidate ranking under partial evaluation.

First, we estimate fitness on a fixed subset of 50 cases per mutant, which provides a practical trade-off between computational cost and estimator variance. To mitigate the noise introduced by subset-based evaluation, we use a \textit{structured batch composition} instead of purely random sampling: each batch consists of 25 random training cases for global coverage, 10 lineage cases inherited from the parent’s evaluation history to maintain comparability between related candidates, and 15 hard cases sampled based on historical failure rates. This composition balances exploration, local consistency, and sustained pressure on difficult examples.

Second, to further stabilize candidate ranking, we introduce \textit{lineage blending}, combining the current mutant accuracy with an exponentially weighted lineage estimate using $\gamma = 0.7$. This strategy reduces sensitivity to subset noise by making selection robust to accidentally easy or difficult batches, preventing spurious promotion of weak mutants and penalization of strong ones. In addition, we apply a conservative rejection rule: a mutant is discarded if its raw score falls below the lineage estimate by more than 0.15, preventing unstable candidates from entering the archive. We selected all hyperparameters based on preliminary experiments and fixed them for all reported runs.

In addition to accuracy, we track the mean number of Expert tokens per case as a proxy for interaction cost and discourage strategies that rely on excessive querying by constraining the maximum number of follow-up questions for baselines and mutants to 8, reflecting typical bounds in real-world clinical consultations~\cite{sviridov20253mdbench}. This constraint helps avoid unnecessarily prolonged interaction patterns observed in prior \mediq-style setups, while preserving sufficient flexibility for informative evidence gathering.


\subsection{Classification of PneumoniaMNIST}
\label{app:pneumoniamnist_details}

\subsubsection{Task Setup}

We formulate medical image classification as \emph{prompt-only} evolutionary optimization for a fixed medical vision-language model. The task is PneumoniaMNIST from the MedMNIST collection, a widely used benchmark for medical image classification. It is a binary pediatric chest X-ray task derived from 5{,}856 chest radiographs, with labels \texttt{normal} and \texttt{pneumonia}~\cite{yang2023medmnist}. We use the MedMNIST+ image versions at all available 2D resolutions, \(28 \times 28\), \(64 \times 64\), \(128 \times 128\), and \(224 \times 224\), and run the experiment independently at each resolution~\cite{doerrich2025medmnistplus}. The official PneumoniaMNIST test split contains 624 images: 234 normal chest radiographs and 390 pneumonia cases.

We evaluate two instruction-tuned multimodal models from the MedGemma family: MedGemma 4B and MedGemma 27B multimodal~\cite{sellergren2025medgemma}. For each model-resolution pair, the underlying VLM, image input, and output schema are fixed during evolution. The evolutionary process can therefore improve performance only by changing the prompt returned by the candidate program, rather than by changing the model, image preprocessing, or evaluation code.

\subsubsection{Candidates and Baselines}

Each candidate is a standalone Python module whose \texttt{entrypoint()} returns a single prompt string. The validator sends this prompt together with a chest X-ray image to the fixed MedGemma model and expects a binary class prediction in a strict JSON object with \texttt{class\_id} and \texttt{class\_name}. This candidate representation makes the search space a set of executable prompt-producing modules while keeping the classification pipeline itself fixed.

The initial population consists of 10 seed prompt programs generated by ChatGPT-5\footnote{\url{https://openai.com/index/introducing-gpt-5/}}. The seed prompts cover several clinically interpretable radiographic instruction styles. Some prompts frame the task as chest-radiograph interpretation and ask the model to consider findings compatible with pneumonia, including focal or lobar air-space opacities, patchy opacities, consolidation, and possible air bronchograms. Other prompts add pediatric chest-radiograph context, such as perihilar or patchy opacities and segmental/lobar consolidation. A separate group of prompts introduces differential-diagnosis guidance for common mimics and confounders, including atelectasis with volume loss or linear opacities, pulmonary edema with more diffuse or bilateral patterns, and technical artifacts such as motion, rotation, and under- or over-exposure.


We use the best seed prompt on the training subset as the \textit{Base} baseline, and compare it with the best evolved prompt produced by MAP-Elites.

\subsubsection{Optimization Objective}

For each image resolution, evolution is run on a fixed balanced subset sampled from the official PneumoniaMNIST training split: 50 normal images and 50 pneumonia images. Candidate fitness is classification accuracy on this 100-image training subset:
\[
\text{fitness} = \frac{\#\text{correct predictions}}{\#\text{evaluated images}}.
\]
We run GigaEvo with a single-island MAP-Elites strategy for 30 generations, using \texttt{gpt-oss-120b} as the mutator model. After evolution, we select two programs by training fitness: the best initial seed prompt and the best evolved descendant prompt. These selected programs are then re-evaluated on the official PneumoniaMNIST test split for the same resolution, containing 234 normal and 390 pneumonia images. Final results are reported from this held-out test split, separately for each MedGemma model and image resolution.
\section{Triage Task Results}
\subsection{Initial Program Pool}
\label{app:initial-programs}
Table~\ref{tab:app-initial-programs} reports the pre-evolution seed pool for the Semigran triage experiment. The programs cover diverse hand-designed strategies, including direct prompting, structured output, chain-of-thought prompting, debate-style prompting, knowledge-base retrieval, severity scoring, and decision-tree logic. Base is the strongest initial program by fitness.

\begin{table}[t]
\centering
\resizebox{\columnwidth}{!}{%
\begin{tabular}{lccccc}
\hline
\textbf{Program} & \textbf{Acc (\%)} & $\mathbf{R_{em}}$ & $\mathbf{R_{ne}}$ & $\mathbf{R_{sc}}$ & \textbf{Fitness} \\
\hline
Base              & $77.3 \pm 1.8$ & 0.60 & 0.88 & 0.84 & 0.747 \\
Decision-Tree     & $75.1 \pm 0.9$ & 0.53 & 0.87 & 0.86 & 0.717 \\
Triage-05-03      & $68.0 \pm 3.0$ & 0.68 & 0.80 & 0.56 & 0.680 \\
Triage-Codex      & $68.9 \pm 2.1$ & 0.60 & 0.74 & 0.73 & 0.676 \\
Json-Structured   & $65.6 \pm 2.8$ & 0.55 & 0.92 & 0.50 & 0.639 \\
Multi-Model       & $62.2 \pm 3.6$ & 0.59 & 0.97 & 0.31 & 0.617 \\
CoT-Single        & $62.9 \pm 4.9$ & 0.54 & 0.85 & 0.50 & 0.616 \\
Debate-Triage     & $64.4 \pm 3.0$ & 0.43 & 0.94 & 0.57 & 0.612 \\
KB-Full-Context   & $62.7 \pm 3.6$ & 0.47 & 0.97 & 0.45 & 0.603 \\
Diff-Dx           & $52.0 \pm 3.0$ & 0.99 & 0.51 & 0.07 & 0.590 \\
Minimal-Prompt    & $59.8 \pm 2.7$ & 0.51 & 0.93 & 0.35 & 0.585 \\
KB-TF-IDF         & $62.4 \pm 2.7$ & 0.32 & 0.90 & 0.65 & 0.579 \\
Severity-Scoring  & $52.7 \pm 2.4$ & 0.65 & 0.00 & 0.93 & 0.546 \\
\hline
\end{tabular}}
\caption{Pre-evolution performance of 13 hand-designed initial programs on the Semigran vignette set (10 runs). Sorted by fitness ($0.85 \cdot \text{Acc} + 0.15 \cdot R_{em}$).}
\label{tab:app-initial-programs}
\end{table}

\subsection{Semigran Vignette Evolution Dynamics}
\label{app:semigran-vignette-evolution-dynamics}
\begin{figure}[htb!]
\centering
\includegraphics[width=\columnwidth]{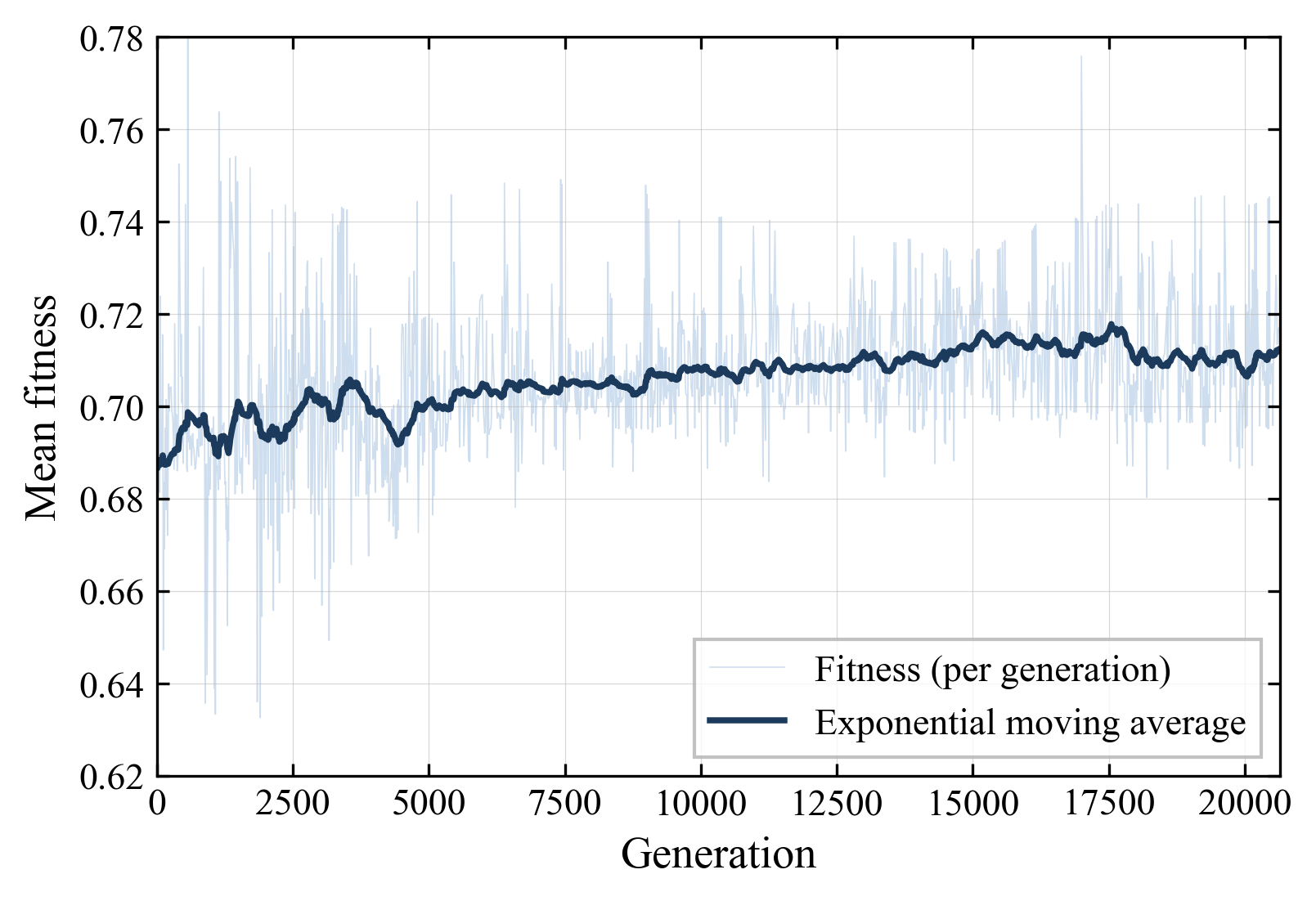}
\caption{Semigran MAP-Elites fitness dynamics over generations.}
\label{fig:triage_fitness}
\end{figure}

Figure~\ref{fig:triage_fitness} shows the MAP-Elites fitness trajectory for the Semigran evolution run. Archive fitness increases steadily across generations, indicating that search continues to discover stronger triage programs rather than saturating early.

\subsection{Cross-Model Transfer}
\label{app:cross-model-transfer}
Table~\ref{tab:transfer} evaluates whether the best Semigran evolved program transfers across unseen LLM backbones without code changes. SG-c1189 preserves high accuracy and emergency recall across all models, while Base remains substantially weaker.

\begin{table}[t]
\centering
\resizebox{\columnwidth}{!}{%
\begin{tabular}{llccccc}
\hline
\textbf{Prog.} & \textbf{Model} & \textbf{Acc (\%)} & $\mathbf{R_{em}}$ & $\mathbf{R_{ne}}$ & $\mathbf{R_{sc}}$ & $\boldsymbol{\Delta}$\textbf{Acc} \\
\hline
Base    & gpt-oss-120b$^\star$  & $77.3 {\pm} 1.8$ & 0.60 & 0.88 & 0.84 & --- \\
\texttt{SG-c1189} & gpt-oss-120b$^\star$  & $87.1 {\pm} 2.0$ & 0.97 & 0.83 & 0.81 & $+9.8$ \\
\hline
Base    & gpt-oss-20b           & $74.2 {\pm} 3.4$ & 0.55 & 0.83 & 0.85 & --- \\
\texttt{SG-c1189} & gpt-oss-20b           & $\mathbf{87.3 {\pm} 2.6}$ & \textbf{0.98} & 0.81 & 0.83 & $+13.1$ \\
\hline
Base    & Qwen-3.5-122B          & $74.7 {\pm} 3.0$ & 0.57 & 0.84 & 0.83 & --- \\
\texttt{SG-c1189} & Qwen-3.5-122B          & $86.9 {\pm} 2.5$ & \textbf{0.99} & 0.79 & 0.83 & $+12.2$ \\
\hline
Base    & Qwen-3.5-27B           & $74.9 {\pm} 2.1$ & 0.55 & 0.89 & 0.81 & --- \\
\texttt{SG-c1189} & Qwen-3.5-27B           & $86.2 {\pm} 2.9$ & 0.96 & 0.81 & 0.82 & $+11.3$ \\
\hline
\end{tabular}}
\caption{Cross-model transfer on the Semigran benchmark (10 runs, $^\star$\ denotes used evolution model). The evolved program transfers without modification to three unseen models, maintaining 86--87\% accuracy and $R_{em} \geq 0.96$ throughout.}
\label{tab:transfer}
\end{table}

\subsection{Held-Out Evaluation}
\label{app:held-out-evaluation}
Table~\ref{tab:heldout} evaluates transfer to the independently constructed Levine vignette set. \texttt{SG-c1189} preserves perfect emergency recall but shifts toward conservative overtriage.

\begin{table}[t]
\centering
\small
\begin{tabular}{lcccc}
\hline
\textbf{Program} & \textbf{Acc (\%) } & $\mathbf{R_{em}}$ & $\mathbf{R_{ne}}$ & $\mathbf{R_{sc}}$ \\
\hline
Base              & $\mathbf{82.2 \pm 2.3}$ & 0.91 & \textbf{0.76} & \textbf{0.86} \\
\texttt{SG-c1189} & $75.1 \pm 2.9$ & \textbf{1.00} & 0.57 & 0.85 \\
\hline
GPT-3 (Levine et al.) & 71 & --- & --- & --- \\
\hline
\end{tabular}
\caption{Held-out evaluation on Levine vignettes~\cite{Levine2023.01.30.23285067} ($n{=}48$, 10 runs). GPT-3 result is from Levine et al.}
\label{tab:heldout}
\end{table}

\subsection{Contextual Comparison on MIMIC-derived ESI Prediction}
\label{app:contextual-comparison-on_mimic}
Table~\ref{tab:mimic_esi_gaber} contextualizes our best MIMIC-ESI program against the reference study. The comparison is not direct, since the systems use different model backbones and evaluation splits, but it provides an external calibration point for exact and range accuracy.

\begin{table}[t]
\centering
\small
\begin{tabular}{lcc}
\hline
\textbf{System} & \textbf{Exact (\%)} & \textbf{Range (\%)} \\
\hline
RAG-assisted LLM & 65.75 & 77.15 \\
Claude 3.5 Sonnet & 64.40 & 82.40 \\
Claude 3 Sonnet & 61.65 & 74.55 \\
Claude 3 Haiku & 59.00 & 66.15 \\
\hline
\texttt{MIMIC-023} (ours) & $62.0 \pm 0.4$ & $77.0 \pm 0.4$ \\
\hline
\end{tabular}
\caption{Contextual comparison on MIMIC-derived ESI prediction. Reference values are clinical-user triage-level results from Gaber et al.~\citep{gaber2025llmworkflows}; ours are evaluated on a separate held-out test split.}
\label{tab:mimic_esi_gaber}
\end{table}

\subsection{MIMIC-ESI Evolution Dynamics}
\label{app:mimic-eci-evolution-dynamics}
\begin{figure}[t]
\centering
\includegraphics[width=\columnwidth]{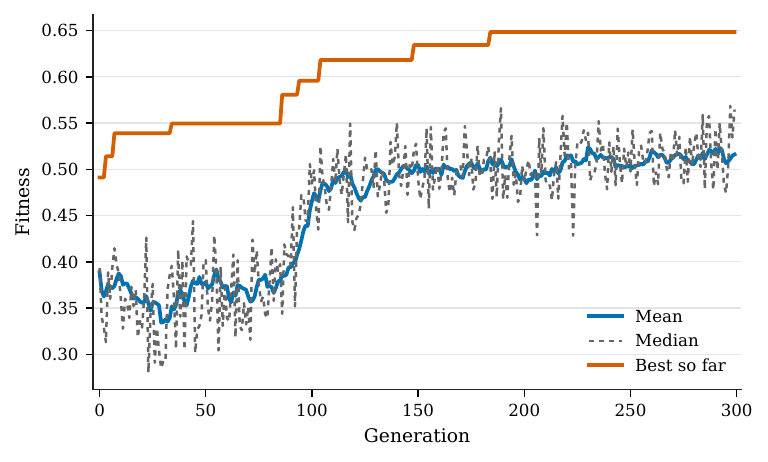}
\caption{MIMIC-ESI evolution dynamics. The plot shows rolling mean archive fitness, median fitness, and the best-so-far frontier over generations.}
\label{fig:mimic_esi_fitness}
\end{figure}

Figure~\ref{fig:mimic_esi_fitness} summarizes the MIMIC-ESI evolutionary run, with the best-so-far frontier showing sustained improvement across generations.

\section{Interactive Consultation Task Results}
\label{app:interactive-consultation-task-results}

\subsection{Performance on \icraftmd{} For Llama-3}
\label{app:perfomance-on-icraft-for-llama-3}
Table~\ref{tab:icraftmd_llama_metrics} evaluates whether the evolved consultation strategy transfers from \imedqa{} to the held-out \icraftmd{} split. Across both Llama-3 models, Evolved achieves the best accuracy while keeping token usage close to the cheapest baseline.

\begin{table}[t]
\centering
\footnotesize
\setlength{\tabcolsep}{3pt}
\begin{tabular*}{\columnwidth}{@{\extracolsep{\fill}}lccc@{}}
\hline
\textbf{Algorithm} & \textbf{Acc (\%) $\uparrow$} & \textbf{Avg. tok. $\downarrow$} & \textbf{Rank $\uparrow$} \\
\hline
\multicolumn{4}{@{}l}{\textbf{Llama-3-8B}} \\
\hline
Scale            & $56.2 \pm 1.2$ & $2727 \pm 10$ & $2+0=2$ \\
Implicit         & $43.6 \pm 4.3$ & $\textbf{154} \pm 2$ & $0+4=4$ \\
Binary           & $\underline{57.4} \pm 0.4$ & $2137 \pm 18$ & $1+1=2$ \\
Numerical cutoff & $56.4 \pm 0.7$ & $3463 \pm 20$ & $3+2=5$ \\
\hline
Evolved          & $\textbf{63.8} \pm 1.1$ & $\underline{156} \pm 13$ & $\textbf{4+3=7}$ \\
\hline
\multicolumn{4}{@{}l}{\textbf{Llama-3-70B}} \\
\hline
Scale            & $75.7 \pm 1.4$ & $2496 \pm 22$ & $3+2=5$ \\
Implicit         & $72.4 \pm 2.7$ & $\textbf{418} \pm 18$ & $0+4=4$ \\
Binary           & $76.2 \pm 2.5$ & $1966 \pm 22$ & $1+0=1$ \\
Numerical cutoff & $\underline{76.4} \pm 1.9$ & $1931 \pm 23$ & $2+1=3$ \\
\hline
Evolved          & $\textbf{78.1} \pm 1.5$ & $\underline{477} \pm 6$ & $\textbf{4+3=7}$ \\
\hline
\end{tabular*}
\caption{\icraftmd{} results on the \mediq{} benchmark with Llama-3 models (3 runs). Rank sums Borda-style scores for accuracy and token usage.}
\label{tab:icraftmd_llama_metrics}
\end{table}

\subsection{Performance On \icraftmd{} For Modern LLM}
\label{app:perfomance-on-icraft-for-modern-llm}
Table~\ref{tab:icraftmd_modern_metrics} evaluates held-out \icraftmd{} transfer with stronger open-weight models. Evolved keeps the highest aggregate rank for both models, matching or approaching the best accuracy while using substantially fewer tokens than explicit cutoff baselines.

\begin{table}[t]
\centering
\footnotesize
\setlength{\tabcolsep}{3pt}
\begin{tabular*}{\columnwidth}{@{\extracolsep{\fill}}lccc@{}}
\hline
\textbf{Algorithm} & \textbf{Acc (\%) $\uparrow$} & \textbf{Avg. tok. $\downarrow$} & \textbf{Rank $\uparrow$} \\
\hline
\multicolumn{4}{@{}l}{\textbf{Qwen-3.5-27B}} \\
\hline
Scale            & $85.1 \pm 2.1$ & $1710 \pm 33$ & $3+0=3$ \\
Implicit         & $62.4 \pm 4.1$ & $\textbf{625} \pm 54$ & $0+4=4$ \\
Binary           & $85.5 \pm 0.8$ & $1753 \pm 24$ & $1+2=3$ \\
Numerical cutoff & $\underline{88.1} \pm 0.8$ & $2214 \pm 7$ & $2+1=3$ \\
\hline
Evolved          & $\textbf{88.3} \pm 3.3$ & $\underline{789} \pm 22$ & $\textbf{4+3=7}$ \\
\hline
\multicolumn{4}{@{}l}{\textbf{Gemma-4-31B}} \\
\hline
Scale            & $88.5 \pm 2.9$ & $1065 \pm 29$ & $4+0=4$ \\
Implicit         & $80.7 \pm 1.2$ & $\underline{294} \pm 22$ & $0+3=3$ \\
Binary           & $89.0 \pm 0.8$ & $844 \pm 8$ & $1+1=2$ \\
Numerical cutoff & $\textbf{89.8} \pm 1.6$ & $1554 \pm 12$ & $2+2=4$ \\
\hline
Evolved          & $\underline{89.5} \pm 0.8$ & $\textbf{46} \pm 2$ & $\textbf{3+4=7}$ \\
\hline
\end{tabular*}
\caption{\icraftmd{} results on the \mediq{} benchmark with modern LLMs (3 runs). Rank sums Borda-style scores for accuracy and token usage.}
\label{tab:icraftmd_modern_metrics}
\end{table}

\subsection{Full \imedqa{} Results}
\label{app:mediq_full_imedqa}

\begin{table}[t]
\centering
\footnotesize
\setlength{\tabcolsep}{3pt}
\begin{tabular*}{\columnwidth}{@{\extracolsep{\fill}}lccc@{}}
\hline
\textbf{Algorithm} & \textbf{Acc (\%) $\uparrow$} & \textbf{Avg. tok. $\downarrow$} & \textbf{Rank $\uparrow$} \\
\hline
\multicolumn{4}{@{}l}{\textbf{Llama-3-8B}} \\
\hline
Binary            & $43.6 \pm 1.0$ & $1671 \pm 2$ & -- \\
Implicit          & $43.5 \pm 0.2$ & $339 \pm 8$  & -- \\
Numerical cutoff  & $45.1 \pm 0.6$ & $2788 \pm 7$ & -- \\
Scale             & $44.2 \pm 0.2$ & $2193 \pm 10$ & -- \\
Best from \mediq{} & $45.8 \pm 1.4$ & -- & -- \\
Evolved           & $\textbf{48.2} \pm 0.4$ & $\textbf{289} \pm 2$ & -- \\
\hline
\multicolumn{4}{@{}l}{\textbf{Llama-3-70B}} \\
\hline
Binary            & $58.6 \pm 0.3$ & $1585 \pm 1$ & -- \\
Implicit          & $57.5 \pm 0.7$ & $813 \pm 14$ & -- \\
Numerical cutoff  & $58.4 \pm 1.7$ & $1625 \pm 3$ & -- \\
Scale             & $57.5 \pm 0.5$ & $1627 \pm 4$ & -- \\
Best from \mediq{} & $60.9 \pm 1.4$ & -- & -- \\
Evolved           & $\textbf{62.2} \pm 0.3$ & $\textbf{514} \pm 1$ & -- \\
\hline
\multicolumn{4}{@{}l}{\textbf{Qwen-3.5-27B}} \\
\hline
Binary            & $70.1 \pm 0.7$ & $1727 \pm 16$ & $1+2=3$ \\
Implicit          & $65.9 \pm 0.9$ & $\textbf{707} \pm 6$ & $0+4=4$ \\
Numerical cutoff  & $\underline{71.1} \pm 1.1$ & $2100 \pm 10$ & $3+0=3$ \\
Scale             & $70.3 \pm 0.4$ & $1860 \pm 21$ & $2+1=3$ \\
Evolved           & $\textbf{73.6} \pm 0.6$ & $\underline{961} \pm 8$ & $\textbf{4+3=7}$ \\
\hline
\multicolumn{4}{@{}l}{\textbf{Gemma-4-31B}} \\
\hline
Binary            & $74.8 \pm 0.5$ & $933 \pm 7$ & $1+2=3$ \\
Implicit          & $72.1 \pm 0.6$ & $\underline{417} \pm 4$ & $0+3=3$ \\
Numerical cutoff  & $\textbf{76.4} \pm 0.4$ & $1412 \pm 1$ & $4+0=4$ \\
Scale             & $75.1 \pm 0.8$ & $1132 \pm 6$ & $2+1=3$ \\
Evolved           & $\underline{75.6} \pm 0.5$ & $\textbf{104} \pm 3$ & $\textbf{3+4=7}$ \\
\hline
\end{tabular*}
\caption{Full per-baseline \imedqa{} results on the \mediq{} benchmark. Results are averaged over three runs. Bold indicates the best result and underline indicates the second-best result within each model block. Rank denotes the Borda-style aggregate over accuracy and average token usage where reported; it is omitted for Llama-3 because the original \mediq{} best-accuracy reference does not report token usage.}
\label{tab:mediq_full_imedqa}
\end{table}

\subsection{Ablation Study}
\label{app:inter-cons-ablation-study}
Table~\ref{tab:mediq_ablation} isolates the contribution of the two fitness-stabilization components used during interactive-consultation evolution. Removing structured batches and lineage blending progressively reduces accuracy, indicating that both help stabilize candidate selection under partial evaluation.

\begin{table}[h]
\centering
\small
\begin{tabular}{lc}
\hline
\textbf{Method} & \textbf{Acc (\%) $\uparrow$} \\
\hline
Full configuration & 48.2 $\pm$ 0.4 \\
\quad w/o structured batch composition & 44.8 $\pm$ 0.7 \\
\quad \quad w/o lineage blending & 43.0 $\pm$ 0.7 \\
\hline
\end{tabular}
\caption{Ablation study of fitness stabilization components on the \imedqa{} test subset of the \mediq{} benchmark on Llama-3-8B (3 runs).}
\label{tab:mediq_ablation}
\end{table}
\section{Extended Qualitative Analysis}
\label{app:extended_qualitative_analysis}

This appendix provides the extended qualitative analysis summarized in Section~\ref{sec:qualitative_results}. We focus on the mechanisms that are difficult to capture with aggregate metrics alone: how evolved programs shift decision boundaries, how they use or avoid external context, how they allocate interaction budget, and how they translate clinical cues into executable prompt or program logic. Representative code excerpts are provided in Appendix~\ref{app:qualitative_snippets}.

\subsection{Triage}
\label{app:extended_qualitative_triage}

\paragraph{Semigran vignette triage.}
The Semigran experiment provides a compact example of evolution changing the operating point of a triage system rather than merely improving formatting. The strongest initial program, \textsc{Base}, already recognizes non-emergency and self-care cases reasonably well, but its emergency recall is low ($R_{\mathrm{em}}{=}0.60$). This is the clinically undesirable direction of error: the system is relatively specific, but it misses too many emergencies. The best evolved program, \texttt{SG-c1189}, shifts this boundary substantially, reaching $R_{\mathrm{em}}{=}0.97$ while preserving useful non-emergency and self-care recall ($R_{\mathrm{ne}}{=}0.83$, $R_{\mathrm{sc}}{=}0.81$). This differs from a trivial conservative solution. MedAsk and several LLM baselines also achieve high emergency recall, but typically lose more self-care specificity, whereas \texttt{SG-c1189} keeps both emergency sensitivity and self-care recognition high.

A second qualitative observation is that the best Semigran solution does not rely on the explicit triage knowledge base. Although several initial programs inject ESI, CTAS, or START reference material, the evolved winner is a compact prompt-based wrapper. In this setting, evolution appears to improve decision framing and class-boundary calibration more than it benefits from additional retrieved context. This is consistent with the knowledge-base ablation: dedicated KB baselines are weaker than both \textsc{Base} and \texttt{SG-c1189}. The result suggests that, for short standardized vignettes, external context can be less useful than a well-calibrated instruction that elicits the model's latent triage knowledge.

The Levine evaluation reveals the cost of this Semigran specialization. \texttt{SG-c1189} transfers its safety behavior---it reaches perfect emergency recall on Levine---but overall accuracy drops because many non-emergency cases are overtriaged. This is a clinically safer failure mode than undertriage, but it shows that optimizing on a small vignette set can still overfit the boundary between \texttt{em} and \texttt{ne}. In contrast, \textsc{Base} generalizes more uniformly to Levine, even though it is weaker on Semigran. Thus, the Semigran result is best interpreted as successful safety-oriented specialization, not as evidence that a single evolved vignette wrapper has solved triage robustness.

\paragraph{MIMIC-ESI triage.}
The MIMIC-ESI experiment reveals two qualitatively different kinds of evolved behavior. The strongest held-out program, \texttt{MIMIC-023}, combines local ESI reference retrieval with structured voting and resource-sensitive reasoning, yielding the best exact accuracy and overall fitness on the held-out MIMIC split. A second family, represented by \texttt{MIMIC-011}, exposes a different point on the search landscape: it is much more conservative, achieving high ESI-1 recall and low undertriage on MIMIC, but at the cost of lower exact accuracy. This kind of candidate is clinically interesting because it optimizes the safety-weighted objective without necessarily optimizing the same operating point that would be preferred for deployment.

The external transfer evaluation shows why this distinction matters. When mapped to Semigran and Levine, \texttt{MIMIC-011} collapses to an all-emergency classifier: it predicts \texttt{em} for all 45 Semigran cases and all 48 Levine cases. Inspecting the evolved code reveals the mechanism. The program contains an early high-acuity rule that returns ESI~1 before any LLM call when systolic blood pressure appears below 90. However, for vignette datasets without structured vitals, the parser returns an empty dictionary and the code evaluates missing SBP with a default value of zero:
\begin{quote}
\small
\texttt{vitals.get("SBP", 0) < 90}
\end{quote}
Thus, absent vitals are interpreted as critical hypotension. This single default value turns a safety check that is useful in the MIMIC setting into a degenerate shortcut under distribution shift.

We view this as a useful failure mode rather than merely a bad candidate. It shows that evolutionary search can discover clinically plausible safety heuristics, but also that such heuristics need robustness constraints around missingness and dataset shift. In triage, where undertriage and overtriage have asymmetric costs, an objective that rewards conservative behavior can produce programs that look attractive on safety metrics while losing specificity out of distribution. This motivates reporting not only exact accuracy and range accuracy, but also prediction distributions, undertriage rates, and external transfer behavior.

\subsection{Interactive Consultation}
\label{app:extended_qualitative_mediq}

For the interactive consultation task, the evolved programs reveal several qualitatively distinct mechanisms for improving the accuracy--cost trade-off. These mechanisms affect different parts of the consultation policy: commitment decisions, fallback behavior, question selection, evidence tracking, and final forced choice. Representative code excerpts for this task are provided in Appendix~\ref{app:qualitative_snippets_mediq}.

\paragraph{Vote-based confidence without an extra scoring call.}
The baseline strategies expose a trade-off between cost and reliability. Explicit confidence-based baselines, such as numerical or Likert-scale cutoffs, make an additional LLM call to estimate a confidence score, while implicit and binary baselines are cheaper but commit whenever any majority vote appears, relying on a simple voting rule. The evolved program for Llama-3-8B keeps the single-pass implicit format, without a separately prompted confidence score, but derives a numeric confidence estimate from the vote distribution itself. It requires a supermajority threshold ($\geq 0.75$), meaning that all three sampled answers must agree before the program commits to a final answer. This strategy preserves the low cost of implicit abstention while adding a simple reliability check before the final answer.

\paragraph{Utility-based question selection.}
The baseline strategies select follow-up questions from self-consistency samples either randomly or by maximum vote frequency, without estimating their expected clinical utility. The Llama-3-70B evolved program introduces a lightweight utility score based on clinically informative attributes, such as onset, duration, severity, risk factors, and family history. This score is combined with vote frequency when choosing the next question, shifting the policy from purely frequency-based selection toward high-yield information gathering. As a result, the agent prioritizes questions that are more likely to reduce diagnostic uncertainty.

\paragraph{Evidence-balanced hypothesis probing.}
The baseline strategies generate follow-up questions without explicitly tracking how much evidence has been collected for each answer option. As a result, fallback questions are either generic or selected from candidate questions without considering which hypotheses remain underexplored. The Qwen-3.5-27B evolved program introduces option-level evidence tracking: it analyzes the interaction history, estimates which answer option has the least accumulated support, and generates a fallback question targeting that option. This implements an evidence-balancing policy, where the agent probes the weakest part of the current hypothesis space instead of asking a generic follow-up question. Such behavior is directly relevant for multiple-choice clinical reasoning, where the goal is to distinguish between competing diagnoses or treatment options.

\paragraph{Selective self-consistency at forced commitment.}
The baseline strategies make a single LLM call when the interaction budget is exhausted and the agent is forced to return a final answer. The Gemma-4-31B evolved program changes this final step by applying self-consistency only at forced commitment: it samples multiple final answers and returns the majority choice. This is a targeted use of additional computation, since the final forced decision is made under the highest uncertainty and directly determines task accuracy. The mechanism therefore reduces variance at the most critical point of the dialogue without increasing sampling throughout the full interaction.

Overall, these qualitative findings show that evolution discovers program-level mechanisms rather than only local prompt edits. The learned policies improve when to answer, what to ask, how to avoid wasted turns, and how to stabilize final decisions, providing a plausible explanation for the strong quantitative accuracy--cost trade-off.

\subsection{Classification of PneumoniaMNIST}
\label{app:extended_qualitative_pneumonia}

For PneumoniaMNIST, evolution operates at the prompt-program level: every candidate is a small Python module whose \texttt{entrypoint()} returns the prompt sent to the fixed MedGemma VLM. The strongest initial programs were already competent format controllers. For 27B, the selected \textsc{Base} prompts were often minimal JSON-only classifiers; for 4B, the selected \textsc{Base} prompts usually added a simple sensitive rule, asking the model to choose pneumonia when it sees a convincing abnormal opacity or infiltrate. The evolved programs preserve this strict output contract, but change the clinical framing and the decision procedure used before the JSON answer is produced.

The most consistent qualitative change is a shift from label-first prompting to finding-oriented image assessment. Evolved prompts introduce a concrete radiology role, often specialized to pediatric AP/PA chest radiographs, and ask the model to inspect both lung fields for focal or lobar air-space opacity, consolidation, patchy or perihilar infiltrates, interstitial patterns, air bronchograms, pleural effusion, and left--right asymmetry. The MedGemma-27B $224{\times}224$ final program in Appendix~\ref{app:qualitative_snippets_pneumonia} is representative: it wraps a strict JSON output contract around an explicit checklist and a deterministic rule that maps any observed pneumonia-compatible finding to class~1. This helps explain why gains persist even though the model, image preprocessing, label set, and parser are fixed.

Evolution also adjusts the operating point differently across resolutions and model sizes. Several 27B final prompts favor pneumonia under ambiguity, which is useful for increasing sensitivity when subtle opacities are present. By contrast, the 4B $28{\times}28$ final prompt becomes more conservative: it asks for at least two separate opacities or a single opacity involving more than about 10\% of a lung field, and defaults to normal under uncertainty. This suggests that the search is not merely adding medical vocabulary; it is tuning decision thresholds to the error profile of a particular model-resolution pair.

Comparing \textsc{Base} and Final prompts suggests two sources of improvement. Many ingredients are recombinations of information already present in the task description and seed prompts: strict JSON formatting, the binary label schema, pediatric AP/PA chest-X-ray framing, focal or lobar air-space opacity, patchy and interstitial opacities, air bronchograms, pleural effusion, common confounders, and the general idea of thresholding the decision boundary. Evolution makes these ingredients more operational by choosing which findings to emphasize, ordering them into an internal checklist, and pairing them with a concrete decision rule. Other details appear to be prompt-level elaborations introduced during mutation and then retained by selection, such as approximate opacity-area thresholds, a minimum perihilar-opacity size, costophrenic-angle blunting, and model-specific uncertainty defaults. In this sense, evolution acts less like unconstrained invention and more like clinically informed recombination plus selective sharpening: it turns task-level medical hints into executable prompt programs whose thresholds are tuned to a particular model-resolution pair.
\section{Representative Evolved Program Snippets}
\label{app:qualitative_snippets}

\subsection{Triage}
\label{app:qualitative_snippets_triage}

\appparagraph{Semigran: urgency-biased consensus without explicit KB context.}
\begin{lstlisting}[style=appendixpython]
_MAX_QUERIES = 30
_BATCH_SIZE = 3
_CONFIDENCE_THRESHOLD = 0.60
_URGENCY_PRIORITY = {"em": 3, "ne": 2, "sc": 1}

def _deterministic_choice(votes: Counter) -> str:
    if not votes:
        return ""
    max_votes = max(votes.values())
    candidates = [c for c, cnt in votes.items() if cnt == max_votes]
    candidates.sort(key=lambda c: _URGENCY_PRIORITY[c], reverse=True)
    return candidates[0]

while total_queries < _MAX_QUERIES:
    current_conf = max(votes.values()) / total_queries
    if current_conf >= _CONFIDENCE_THRESHOLD:
        break

    batch_votes = await _vote_batch(prompt, _BATCH_SIZE)
    votes.update(batch_votes)
    total_queries += _BATCH_SIZE
\end{lstlisting}

\appparagraph{MIMIC-023: retrieval, resource reasoning, and safety-biased voting.}
\begin{lstlisting}[style=appendixpython]
TOP_K = 3
SIMILARITY_CUTOFF = 0.30

def retrieve(text: str) -> str:
    ...
    for idx, _ in ranked:
        sim = _cosine(query_vec, _VECTORS_CACHE[idx], _NORMS_CACHE[idx])
        if sim < SIMILARITY_CUTOFF:
            continue
        parts.append(f"### {_KB_CACHE[idx]['title']}\n{snippet}")
        if len(parts) >= TOP_K:
            break
    return "\n\n".join(parts)

def majority_vote(votes: List[str]) -> str:
    cnt = Counter(votes)
    if cnt["1"] >= 1:
        return "1"
    if cnt["2"] >= 2:
        return "2"
    max_cnt = max(cnt.values())
    tied = [int(v) for v, c in cnt.items() if c == max_cnt]
    return str(min(tied))  # lower ESI number is more urgent

raw_votes = await asyncio.gather(*(one_vote(txt, sys) for sys in systems))
res = await ask_llm_json(txt, RESOURCE_SYSTEM)
res_vote = resource_vote(txt, parse_count(res.get("resource_count")))
valid_votes = [v for v in raw_votes if v is not None] + [res_vote]
pred = majority_vote(valid_votes)
\end{lstlisting}

\appparagraph{MIMIC-011: missing vitals become a brittle safety shortcut.}
\begin{lstlisting}[style=appendixpython]
def high_acuity_rule(patient: dict) -> bool:
    txt = patient_text(patient).lower().replace("-", " ")
    ...
    vitals = _parse_vitals(patient.get("initial_vitals", ""))

    if vitals.get("SBP", 0) < 90:
        return True
    if vitals.get("HR", 0) > 130:
        return True
    if vitals.get("RR", 0) > 30:
        return True
    if vitals.get("SpO2", 100) < 90:
        return True
    return False

async def classify(patient: dict) -> dict:
    if high_acuity_rule(patient):
        return {"predicted_acuity": "1"}
    ...
\end{lstlisting}

\subsection{Interactive Consultation Task}
\label{app:qualitative_snippets_mediq}

\appparagraph{Vote-based confidence without an extra scoring call.}
\begin{lstlisting}[style=appendixpython]
...

if answer_votes:
    chosen_letter = max(set(answer_votes), key=answer_votes.count)
    confidence = answer_votes.count(chosen_letter) / total_iters
    raw = answer_texts.get(chosen_letter, "")
elif question_votes:
    ...

if answer is not None and confidence >= CONFIDENCE_THRESHOLD:
    return {
        "abstain": False,
        "confidence": confidence,
        "messages": messages,
        "letter_choice": answer,
        "atomic_question": None,
    }
\end{lstlisting}

\appparagraph{Utility-based question selection.}
\begin{lstlisting}[style=appendixpython]
...

HIGH_VALUE_KEYWORDS = [
    "duration", "onset", "severity", "frequency",
    "risk factor", "family history", "exposure",
    "pain", "fever", "cough", "headache",
]

def _score_question(question: str) -> int:
    lower = question.lower()
    return sum(keyword in lower for keyword in HIGH_VALUE_KEYWORDS)

scored = {
    question: _score_question(question) + Counter(questions)[question]
    for question in set(questions)
}

best_question = max(scored, key=scored.get)
\end{lstlisting}

\appparagraph{Evidence-balanced hypothesis probing.}
\begin{lstlisting}[style=appendixpython]
def _suggest_discriminative_question(
    patient_state,
    options_dict,
    asked_options,
):
    vote_counts = {option: 0 for option in options_dict}

    for qa in patient_state["interaction_history"]:
        answer = parse_choice(qa.get("answer", ""), options_dict)
        if answer:
            vote_counts[answer] += 1

    remaining = [
        option for option in options_dict
        if option not in asked_options
    ]

    if remaining:
        target = min(remaining, key=lambda o: vote_counts.get(o, 0))
    else:
        target = min(vote_counts, key=vote_counts.get)

    return (
        f"Does the patient exhibit any findings "
        f"that would support diagnosis {target}?"
    )
\end{lstlisting}

\appparagraph{Selective self-consistency at forced commitment.}
\begin{lstlisting}[style=appendixpython]
...

n = PROGRAM_PARAMS["self_consistency"]
responses = get_responses(messages, n=n, **kwargs)

letters = []
for response in responses:
    _, letter = expert_response_choice(
        messages,
        options_dict,
        forced_response=response,
        **kwargs,
    )
    if letter in ("A", "B", "C", "D"):
        letters.append(letter)
...

return max(set(letters), key=letters.count)
\end{lstlisting}

\subsection{Classification of PneumoniaMNIST}
\label{app:qualitative_snippets_pneumonia}

\appparagraph{Checklist-driven prompt program for MedGemma-27B.}
\begin{lstlisting}[style=appendixpython]
# Program implementation
PROMPT = """You are an expert pediatric radiologist.

Task: classify the chest X-ray (pediatric AP/PA view) as normal vs pneumonia.

Think internally, then output ONLY a single-line JSON object with EXACTLY these keys:
{"class_id": <0 or 1>, "class_name": "<normal or pneumonia>"}

Guidelines:
1) Systematically inspect the right and left lung fields for any of the following findings:
   - focal or lobar airspace opacity
   - consolidation
   - patchy infiltrates
   - air bronchograms
   - perihilar opacity >=0.5 cm or diffuse subtle opacities
2) If any such finding is present in either lung, label pneumonia (class_id 1, class_name "pneumonia").
3) If findings are present in both lungs, choose the dominant pathology (the lung with the more extensive opacity) - still pneumonia.
4) If no findings from step 1 are observed, label normal (class_id 0, class_name "normal").
5) If you are uncertain, select the class you deem most likely based on the visual evidence (do NOT default to normal).
6) Do NOT include any explanations, markdown, or extra characters before or after the JSON. The output must be a single line.

"""

def entrypoint():
    return PROMPT
\end{lstlisting}

\clearpage
\twocolumn[\vskip\intextsep
\noindent\begin{minipage}{\textwidth}
\centering
\small
\resizebox{\textwidth}{!}{
\begin{tabular}{lccccc}
\hline
\textbf{Task} &
\textbf{Runs} &
\textbf{Evolver calls/run} &
\textbf{Evolver tokens/call} &
\textbf{Expert calls/eval} &
\textbf{Expert tokens/call} \\
\hline

\makecell[l]{MIMIC-ESI triage\\(estimated)} &
$1$ &
$2{,}101$ &
\makecell[l]{in: $9{,}932/11{,}472/12{,}143$\\out: $2{,}468/2{,}666/3{,}009$} &
$214/347/358$ &
\makecell[l]{in: $327/429/583$\\out: $7/7/7$} \\

Interactive consultation &
$4$ &
$1{,}499/1{,}698/2{,}302$ &
\makecell[l]{in: $6{,}153/7{,}954/10{,}880$\\out: $1{,}063/1{,}295/2{,}884$} &
$266/387/535$ &
$3{,}606/6{,}337/19{,}252$ \\


\makecell[l]{PneumoniaMNIST} &
$24$ &
$240$ &
\makecell[l]{in: $2{,}140/2{,}159/2{,}182$\\out: $826/966/1{,}097$} &
$100$ &
$402/429/440$ \\

\hline
\end{tabular}}
\captionof{table}{Evolution resource consumption across tasks. Evolver statistics describe LLM-guided mutation calls, while expert statistics describe task-model calls used during candidate evaluation. Calls are reported per evolutionary run, and token counts are reported as 25th/50th/75th percentiles per call.}
\label{tab:evolution_resources}
\end{minipage}
\par\vskip\intextsep
]
\section{Evolution Costs}
\label{app:evo_costs}

Table~\ref{tab:evolution_resources} reports the resource consumption of the evolutionary pipelines. For the MIMIC-ESI triage task, a single evolutionary run produced 2{,}101 non-root candidate programs. Each candidate was evaluated on a stratified batch of 96 patient cases, and the final evolved wrappers typically made multiple expert-model calls per case because they used combinations of ESI-level voting, resource-count estimation, retrieval, and safety checks. This leads to a relatively high number of expert calls per evaluation batch, but the individual expert prompts are short compared with dialogue-based tasks.

For the interactive consultation task, a median run corresponds to an estimated API cost of roughly \$2.30--5.18 when using a 27B+ expert model, depending on model choice and provider pricing\footnote{\url{https://computeprices.com/models}}. As a point of reference, fine-tuning introduces GPU-provisioning and training costs that depend on model size, sequence length, and GPU architecture. Even under an efficient setup, prior cost modeling reports \$17.9--32.7 for a single sparse Mixtral-8x7B fine-tuning run, using a 47B-parameter MoE model adapted with QLoRA~\citep{xia2024understanding}. Thus, for low-to-medium run counts, inference-time evolution can be a practical alternative to fine-tuning because it avoids dedicated training infrastructure, although its API-based cost accumulates with repeated searches and large-scale deployment~\citep{klang2024strategy}.

For PneumoniaMNIST, the evolutionary setup is structurally simpler because each candidate is only a prompt-producing Python module. However, evaluation is image-call intensive: each candidate is scored on a fixed balanced subset of 100 training images, so one candidate evaluation requires 100 MedGemma calls. We ran four image resolutions and two MedGemma model sizes, with three repeated evolutionary runs  per model--resolution pair, for a total of 24 evolutionary runs. Each run used 30 generations with 8 mutants per generation, corresponding to 240 evolver calls per run. Compared with the interactive consultation task, PneumoniaMNIST uses much shorter textual contexts. Its cost profile is therefore not driven by long dialogue histories or large textual contexts, but by the multiplicity of independent image evaluations.

\end{document}